\pdfoutput=1
\documentclass[10pt]{article}
\usepackage{times}

\usepackage{graphicx}
\usepackage{float}
\usepackage{flafter}
\usepackage{hyperref}
\usepackage{caption}
\usepackage{lineno}
\usepackage{booktabs}
\usepackage{array}
\usepackage{xcolor}

\usepackage{amssymb}
\newcommand{\cmark}{\checkmark}
\newcommand{\xmark}{$\times$}

\definecolor{defaultcolor}{gray}{0.9}
\newlength\savewidth

\usepackage{amsmath,amsfonts,bm}
\usepackage{graphicx}
\usepackage{grffile}
\graphicspath{{./}{nmed_sections/}}

\definecolor{defaultcolor}{gray}{0.9}

\def\eqref#1{equation~\ref{#1}}

\def\1{\bm{1}}

\DeclareMathAlphabet{\mathsfit}{\encodingdefault}{\sfdefault}{m}{sl}
\SetMathAlphabet{\mathsfit}{bold}{\encodingdefault}{\sfdefault}{bx}{n}

\usepackage{caption}

\newcounter{extfigure}
\renewcommand{\theextfigure}{\arabic{extfigure}}

\usepackage[symbol]{footmisc}
\usepackage[noblocks]{authblk}

\usepackage{booktabs}
\usepackage{graphicx}
\usepackage{multirow}
\usepackage{color, colortbl}
\usepackage{caption}
\usepackage{subcaption}
\usepackage{wrapfig}
\usepackage{makecell}
\usepackage{amsfonts}
\usepackage{amsmath}
\usepackage{booktabs}
\usepackage{xspace}
\usepackage{array}
\usepackage{siunitx}
\usepackage{enumitem}
\usepackage[most]{tcolorbox}
\usepackage[square, numbers, sort&compress]{natbib}
\usepackage[margin=1in]{geometry}
\usepackage[normalem]{ulem}
\usepackage{fancyhdr}
\usepackage{soul}

\newcolumntype{F}[1]{%
    >{\raggedright\arraybackslash\hspace{0pt}}p{#1}}%
\newcolumntype{T}[1]{%
    >{\centering\arraybackslash\hspace{0pt}}p{#1}}%

\usepackage[margin=1in]{geometry}
\usepackage[symbol]{footmisc}
\usepackage{booktabs}
\usepackage{longtable}
\usepackage{pdflscape}
\usepackage{threeparttablex}

\newcommand{\eat}[1]{\ignorespaces}

\usepackage[capitalize]{cleveref}
\crefformat{section}{\S#2#1#3}
\Crefname{figure}{\textbf{Figure}}{}
\Crefname{table}{\textbf{Supplementary Table}}{}
\Crefname{algorithm}{Algorithm}{}
\Crefname{algocf}{Algorithm}{}
\Crefname{equation}{Equation}{}
\crefname{appendix}{Appendix}{}
\title{A radiographic world model for clinical reasoning and evidence generation}

\author{
Suyang Xi$^{1,*}$,
Songtao Hu$^{1,*}$,
Shansong Wang$^{2}$,
Mojtaba Safari$^{2}$,
Luke del Balzo$^{3}$,

Ehsan Ul Karim$^{4}$,
Mingzhe Hu$^{1}$,
Kuo Zhang$^{4}$,
Tonghe Wang$^{5}$,

Ralph R. Weichselbaum$^{2}$
and Xiaofeng Yang$^{1,2,3,\dagger}$

\vspace{0.5em}

$^{1}$Department of Computer Science and Informatics, Emory University, Atlanta, GA, USA.\\
$^{2}$Department of Radiation and Cellular Oncology, The University of Chicago, Chicago, IL, USA.\\
$^{3}$Department of Radiation Oncology and Winship Cancer Institute, Emory University, Atlanta, GA, USA.\\
$^{4}$Department of Radiology and Imaging Science, Emory University, Atlanta, GA, USA.\\
$^{5}$Department of Medical Physics, Memorial Sloan Kettering Cancer Center, New York, NY, USA.
}
\footnotetext{* Equal contribution.}
\footnotetext{$\dagger$ Corresponding author, Email: \textcolor{blue}{xfyang@uchicago.edu}}

\date{}
\begin{document}

 \maketitle
\begin{abstract}

Medical imaging artificial intelligence (AI) is commonly developed as separate mappings from radiographs to diagnostic outputs or from clinical descriptions to generated images, although both arise from the same underlying radiographic state. A world-model formulation instead seeks to learn an internal representation of this state that can support both clinical readout and conditional simulation of radiographic observations. Here we introduce MedDream, a radiographic world model that learns a shared continuous latent state from paired chest radiograph–text observations for diagnostic reasoning and report-conditioned evidence generation. MedDream was pretrained on 2.65 million leakage-controlled chest radiograph–text pairs curated from 4.40 million candidates. Across eight clinical datasets and two independent reader cohorts, MedDream outperformed leading diagnostic and generative comparators. For diagnostic reasoning, MedDream showed strong generalization across disease recognition, label-scarce adaptation, severity assessment, and localization, while MedDream-supported review increased mean resident concordance with independent radiologist consensus from 56.3\% to 63.0\%. For evidence generation, MedDream produced radiographs that preserved clinically relevant pathology and improved downstream performance on held-out real data, with synthetic augmentation increasing external VinDr-CXR macro-AUROC from 76.4\% to 81.4\%. More importantly, conditioning generation on prespecified subgroup performance gaps enabled targeted evidence construction, increasing weighted F1 by 3.1 percentage points in Asian patients, whereas matched-volume unguided augmentation decreased it by 2.3 points. These findings establish radiographic world models as a path toward medical AI that learns clinically meaningful internal states for interpreting, simulating, and constructing evidence for clinical use.

\end{abstract}

\newpage

\section*{Introduction}

Medical imaging is an integral component of the diagnostic process in the modern era, yet rare or atypical findings remain difficult to interpret because comparable clinical examples are scarce and disease-related changes can be confounded by individual anatomical variation~\cite{moor2023foundation,zhang2024generalist,ktena2024generative,wang2025self}. The same evidence gap limits AI in medical imaging: the vast majority of long-tailed clinical cohorts underrepresent rare phenotypes and complex anatomical presentations, thus allowing models to perform well on average, while relying on unstable cues in sparsely represented settings~\cite{xu2024whole,chen2021synthetic}. Medical imaging AI therefore requires two complementary capabilities: learning transferable diagnostic representations from available clinical data and constructing clinically plausible visual examples when corresponding real observations are scarce~\cite{yan2025multimodal}.


Foundation models have begun to address these requirements, but diagnostic representation learning and medical image generation have largely developed as separate capabilities. Large-scale visual pretraining and vision--language alignment have enabled medical imaging models to learn transferable diagnostic representations from images and reports~\cite{tiu2022expert,zhou2023foundation,zhang2024generalist}, supporting disease recognition and abnormality localization across downstream tasks~\cite{lu2024visual,ma2025fully}. These models, however, cannot construct clinically plausible imaging evidence when relevant real examples are scarce. In parallel, diffusion and related generative frameworks have improved the realism and controllability of medical image generation~\cite{bluethgen2025vision,ktena2024generative,konz2024anatomically}, enabling enrichment of rare findings and construction of matched imaging cohorts~\cite{luo2024autoregressive}. Yet generated images are typically guided or evaluated using separately trained image--text models or disease classifiers, so diagnostic and generative learning remain only weakly coupled~\cite{bluethgen2025vision}. Recent unified approaches have begun to narrow this separation~\cite{zhang2026unix,lee2024llm}, but often still rely on distinct representation or optimization pathways rather than a shared representation of the underlying radiographic state.

The main constraint to this coupling is that the two objectives place competing demands on pretraining. Transferable diagnostic representations require preservation of anatomically and pathologically meaningful features, whereas generative learning benefits from stronger corruption that forces recovery of missing anatomy and disease-specific appearance~\cite{li2023mage,li2026dream}. In medical images, mild masking preserves subtle diagnostic cues but provides limited supervision for global anatomical recovery, whereas stronger masking improves generative learning but can obscure localized evidence needed for clinical alignment~\cite{li2026dream,he2022masked,tian2020makes,yao2025reconstruction}. Existing hybrid designs often avoid this tension by separating diagnostic encoders from generative modules or by staging representation learning before generation~\cite{li2023blip,chu2025usp,li2026unified,zhang2026unix}. This separation limits how strongly generative learning can shape the shared radiographic state~\cite{li2023mage,li2025mergevq}, leaving unresolved whether diagnostic readout and clinically faithful generation can be jointly learned within a single visual pathway.

\begin{figure}[p]
\centering
\includegraphics[width=0.98\textwidth]{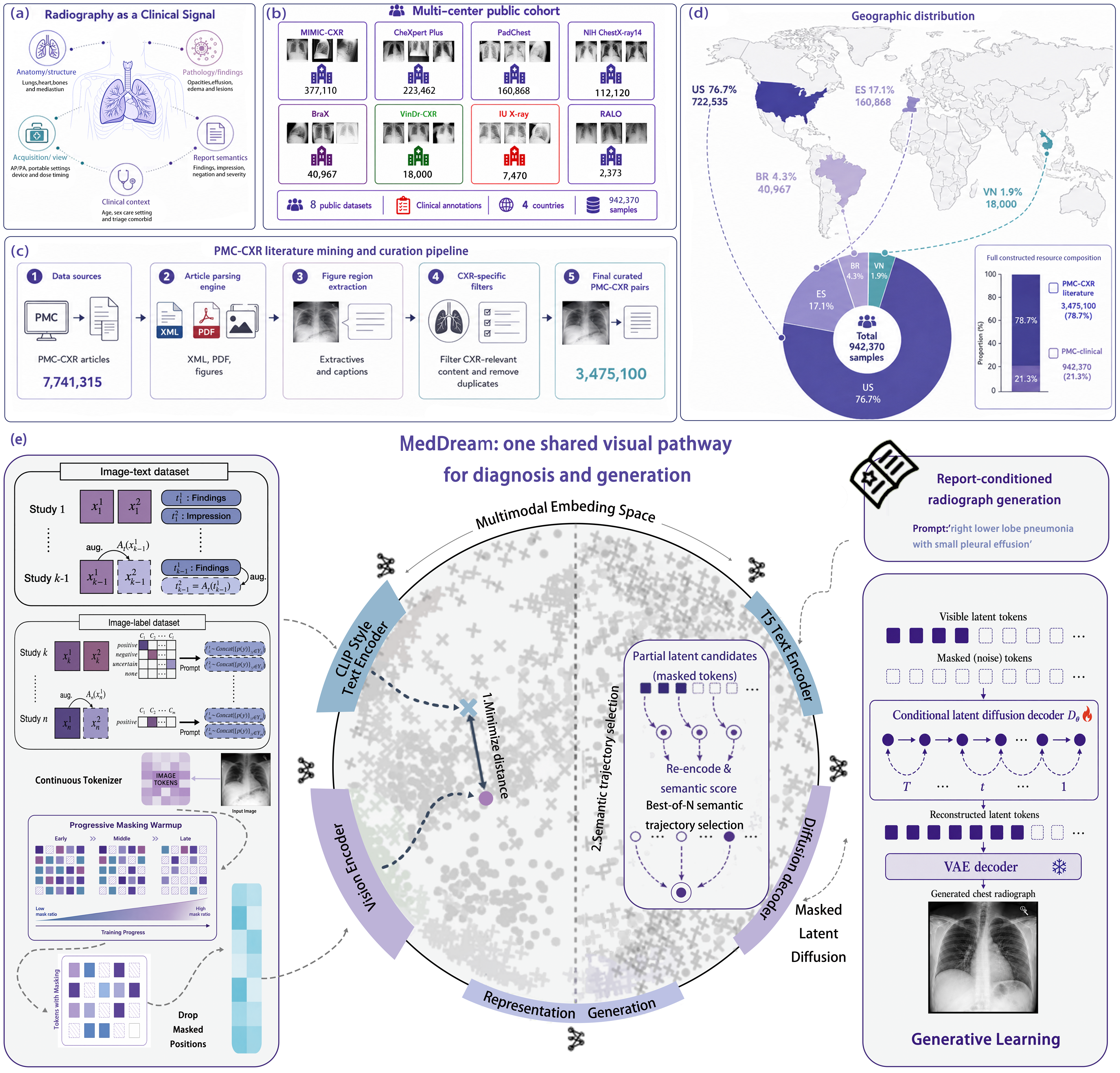}
\caption{
\textbf{MedDream jointly learns a shared medical visual representation for diagnostic adaptation and report-conditioned evidence construction.}
\textbf{a}, Chest radiographs encode anatomical, pathological, acquisition-related, semantic and clinical-contextual information.
\textbf{b}, Public chest-radiography data resources collected during corpus construction. Seven repositories contributed image--text pairs to the candidate corpus, whereas RALO was used only as an independent evaluation resource and did not contribute to MedDream pretraining.
\textbf{c}, Literature-mining pipeline for extracting and curating  candidate chest-radiograph image--caption pairs from open-access biomedical articles.
\textbf{d}, Geographic distribution and source composition of the constructed chest-radiography image--text resource, including literature-mined PMC-CXR pairs and curated clinical image--text pairs before filtering to obtain the final leakage-controlled pretraining corpus.
\textbf{e}, MedDream jointly optimizes clinical image--text alignment and masked latent generation through the same continuous visual pathway. Both objectives update the shared image encoder, yielding a representation that supports frozen diagnostic adaptation and report-conditioned evidence construction.
}
\label{fig:data1}
\end{figure}

Here, we introduce MedDream, a radiographic world model in which diagnostic reasoning and report-conditioned image generation share one pretrained visual pathway (Fig.~\ref{fig:data1}). Rather than pairing a diagnostic encoder with a separately optimized text-to-image generator, MedDream learns both functions over shared continuous visual tokens. The model jointly optimizes clinical image--text alignment and masked latent generation, so that the representation used for frozen diagnostic transfer also conditions image synthesis. A progressive masking curriculum exposes the model to complementary views of radiographic anatomy: low masking preserves visible disease cues for semantic alignment, whereas heavier masking drives recovery of anatomy and pathology for generation. In this way, MedDream couples diagnostic readout and evidence generation within a single visual pathway, enabling clinical interpretation and the construction of synthetic diagnostic evidence when comparable real observations are scarce.


To systematically evaluate MedDream, we used chest radiography as a clinically important test bed, given its widespread use in routine care and the availability of extensive image--text resources and established diagnostic benchmarks~\cite{johnson2019mimic,irvin2019chexpert,nguyen2022vindr}. We curated approximately 4.4 million candidate pairs from PubMed Central biomedical image--caption data and public chest-radiography image--report datasets~\cite{zhang2025multimodal,johnson2019mimic}, from which we derived a leakage-controlled subset for MedDream. We evaluated MedDream along two complementary tracks reflecting its world-model design. The diagnostic-readout track assessed frozen-representation transfer across held-out benchmarks and cohorts excluded entirely from pretraining, covering disease recognition under common, expanded-label, long-tailed, and label-scarce settings, together with severity assessment and spatial localization~\cite{ma2025fully,tiu2022expert,irvin2019chexpert,wang2017chestx,nguyen2022vindr}. The evidence-generation track evaluated report-conditioned generation using distributional and pathology-consistency metrics, complemented by blinded expert review~\cite{bluethgen2025vision,ji2026generative,ktena2024generative,wang2025self}. Downstream studies further examined whether MedDream-generated images could improve learning from limited real data and enable metadata-guided subgroup stress testing and rebalancing on held-out real cohorts~\cite{luo2025fairdiffusion,chen2021synthetic,zhang2025generative}. Together, these experiments test whether a shared radiographic state can support both diagnostic readout and clinically grounded evidence generation when relevant real observations are scarce, long-tailed, or unevenly distributed.

\clearpage
\clearpage

\section*{Results}

\subsection*{MedDream preserves diagnostic transfer across held-out and label-scarce settings.}

\paragraph{In-domain thoracic disease recognition.}

\begin{figure}[!htbp]
\centering
\includegraphics[width=\textwidth]{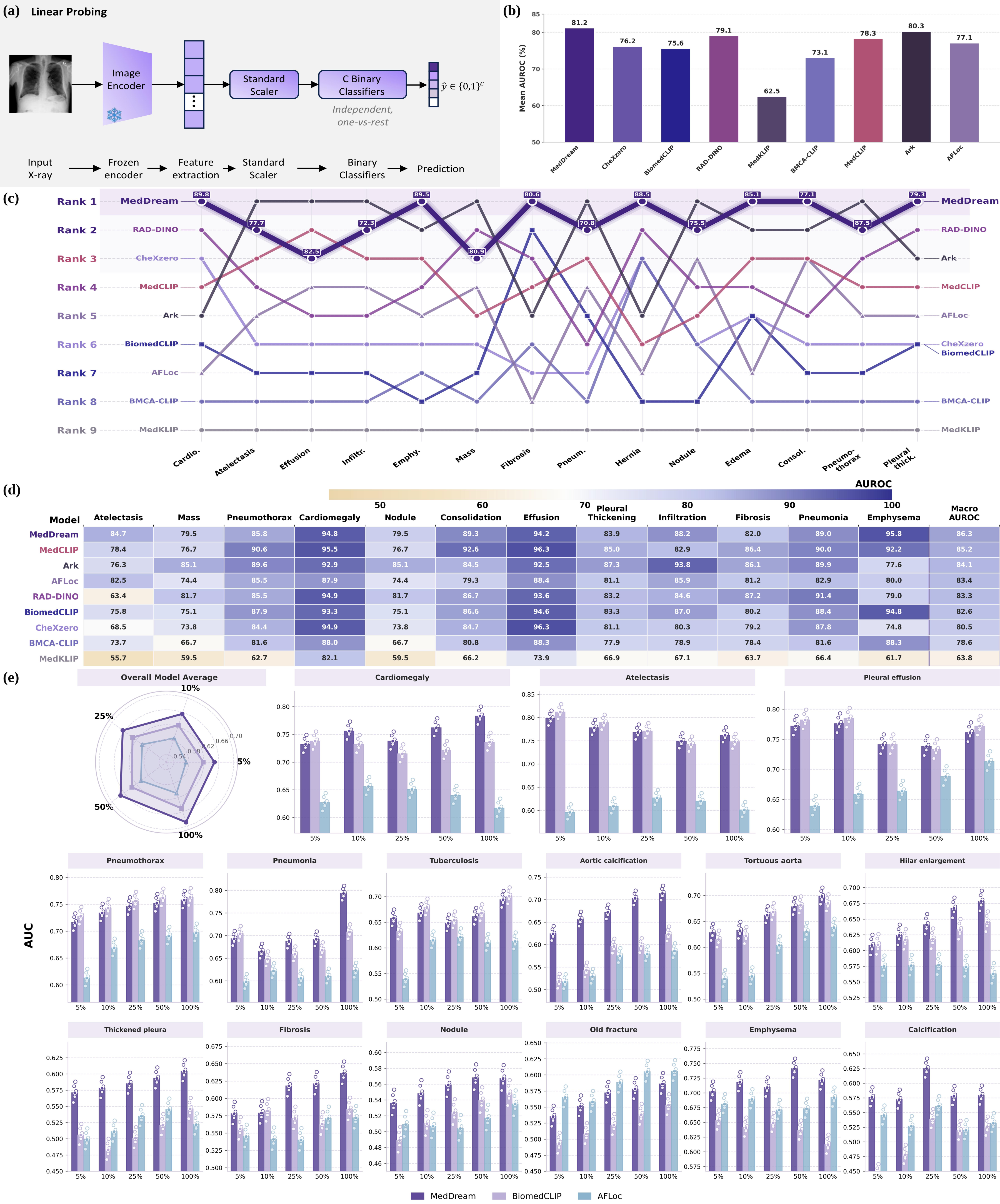}
\caption{\textbf{Diagnostic representation transfer across clinical settings.}
\textbf{a,} Frozen-encoder linear-probing protocol. Chest radiographs were encoded by a fixed image encoder, standardized and passed to independent one-versus-rest binary classifiers.
\textbf{b,} Mean AUROC on the NIH ChestX-ray14 hold-out test set for 14 common thoracic findings. Bars show model-level mean AUROC.
\textbf{c,} Disease-wise rank trajectories on NIH ChestX-ray14 across evaluated thoracic findings. Lower rank indicates better disease-level performance.
\textbf{d,} Disease-wise AUROC heat map on VinDr-CXR comparing MedDream with CheXzero, BiomedCLIP, RAD-DINO, MedKLIP, BMCA-CLIP, MedCLIP, Ark+ and AFLoc.
\textbf{e,} Label-efficiency evaluation on ChestDR. Linear probes were trained with 5\%, 10\%, 25\%, 50\% or 100\% of the labelled training set and evaluated by AUC for overall and disease-wise performance.}
\label{fig:diagnostic_transfer}
\end{figure}

We first tested whether coupling diagnostic representation learning with generative modeling preserved performance on standard thoracic disease recognition. NIH ChestX-ray14 was treated as an in-domain held-out benchmark: its model-development partition contributed to pretraining, whereas the official test partition was excluded from pretraining, model selection, and hyperparameter tuning. Under the same frozen-encoder linear-probing protocol across representative chest-radiography foundation and medical vision-language models~\cite{ma2025fully,wang2022medclip,zhang2023biomedclip,perez2025exploring,tiu2022expert,yang2026multimodal,lozano2025biomedica,wu2023medklip}, MedDream achieved a mean AUROC of 81.2\% across 14 findings (95\% CI, 80.4--82.0\%), exceeding Ark+ at 80.3\% (adjusted $P=0.007$) and MedCLIP at 78.3\% (Fig.~\ref{fig:diagnostic_transfer}b). MedDream also ranked favorably across individual findings (Fig.~\ref{fig:diagnostic_transfer}c), indicating that joint diagnostic-generative pretraining preserved strong frozen-representation performance on held-out data from the same imaging source.

\paragraph{Transfer to a broader diagnostic label space.}

Clinical imaging datasets are often incompletely labeled, and local diagnostic taxonomies may expand as institutions refine reporting practices and introduce more fine-grained findings. We therefore examined whether the frozen MedDream representation transferred to the broader label space of VinDr-CXR, which was entirely excluded from MedDream pretraining. Across 23 diagnostic endpoints spanning global thoracic abnormalities and fine-grained local findings, MedDream achieved a mean AUROC of 86.3\% (95\% CI, 85.3--87.3\%), exceeding Ark+, the strongest comparator, at 85.2\% (adjusted $P=0.004$), as well as the remaining baselines (Fig.~\ref{fig:diagnostic_transfer}d). Performance remained strong for representative lower-prevalence findings. For other lesion ($n=94$), calcification ($n=194$), atelectasis ($n=86$) and interstitial lung disease ($n=221$), MedDream achieved AUROCs of 85.7\% (95\% CI, 81.0--90.4\%), 82.4\% (95\% CI, 78.2--86.6\%), 86.5\% (95\% CI, 81.6--91.4\%) and 85.1\% (95\% CI, 80.9--89.3\%), respectively (Extended Data Fig.~\ref{fig:extended_diagnostic_adaptation}a). MedDream also outperformed MedCLIP and AFLoc across all four endpoints; complete effect sizes, precision--recall analyses and multiplicity-adjusted comparisons are reported in Extended Data Fig.~\ref{fig:extended_diagnostic_adaptation} and Supplementary Table~\ref{tab:diagnostic_transfer_statistics}. These findings indicate that MedDream retained diagnostic utility on a pretraining-excluded dataset with a broader and more fine-grained label space.

\paragraph{Preservation of diagnostic performance under long-tailed label scarcity.}
Chest radiographic diagnosis is inherently long-tailed: common findings dominate local datasets, whereas many clinically important abnormalities have few labeled examples. We evaluated this setting on ChestDR, a pretraining-excluded thoracic disease dataset with 19 labels and a long-tailed class distribution. Linear probes were trained using 5\%, 10\%, 25\%, 50\% or 100\% of the available labeled training data (Fig.~\ref{fig:diagnostic_transfer}e). Under the most label-scarce 5\% setting, MedDream achieved a mean AUROC of 64.8\% (95\% CI, 63.0--66.6\%), compared with 61.9\% for BiomedCLIP and 57.2\% for AFLoc (adjusted $P<0.001$). This pattern was maintained when all labeled data were used: MedDream reached 68.9\% (95\% CI, 67.4--70.4\%), compared with 65.0\% and 60.7\%, respectively (adjusted $P<0.001$). A marked separation was also observed for clinically relevant tail findings under low-label training. At the 5\% label fraction, MedDream achieved 62.4\% AUROC for aortic calcification (95\% CI, 58.5--66.3\%), compared with 51.8\% for BiomedCLIP and 51.7\% for AFLoc (adjusted $P<0.001$). For emphysema, MedDream reached 70.2\% AUROC (95\% CI, 66.9--73.5\%), compared with 65.3\% and 68.1\%, respectively (adjusted $P=0.009$). These findings indicate that MedDream retained label-efficient transfer on a pretraining-excluded, long-tailed cohort, particularly for rare abnormalities.


\subsection*{MedDream preserves severity and spatial evidence for clinically richer interpretation.}

\paragraph{Severity prediction.}
Clinical deployment requires estimating disease extent, rather than only detecting disease presence in binary, because severity can influence monitoring intensity, triage and escalation of care. We evaluated this capability on the held-out RALO dataset, where chest radiographs were assigned to four imbalanced severity levels: none or trace, mild, moderate, and severe. Under the same frozen-encoder linear-probing protocol, MedDream achieved a macro F1 score of 47.2\% (95\% CI, 44.2--50.2\%) and a macro AUROC of 74.8\% (95\% CI, 72.3--77.3\%), exceeding BiomedCLIP, the strongest comparator, which reached 44.0\% and 74.0\%, respectively (adjusted $P = 0.035$ and $0.024$ for macro F1 and macro AUROC, respectively; Extended Data Fig.~\ref{fig:ralo_severity}). Severity-wise analysis showed that MedDream achieved the highest F1 across all four severity levels, including mild disease (48.7\%) and severe disease (47.7\%), two categories in which misestimation may affect monitoring or escalation of care. Confidence intervals and multiplicity-adjusted comparisons are reported in Supplementary Table~\ref{tab:diagnostic_transfer_statistics}.

\begin{figure}[!htbp]
\centering
\includegraphics[width=\textwidth]{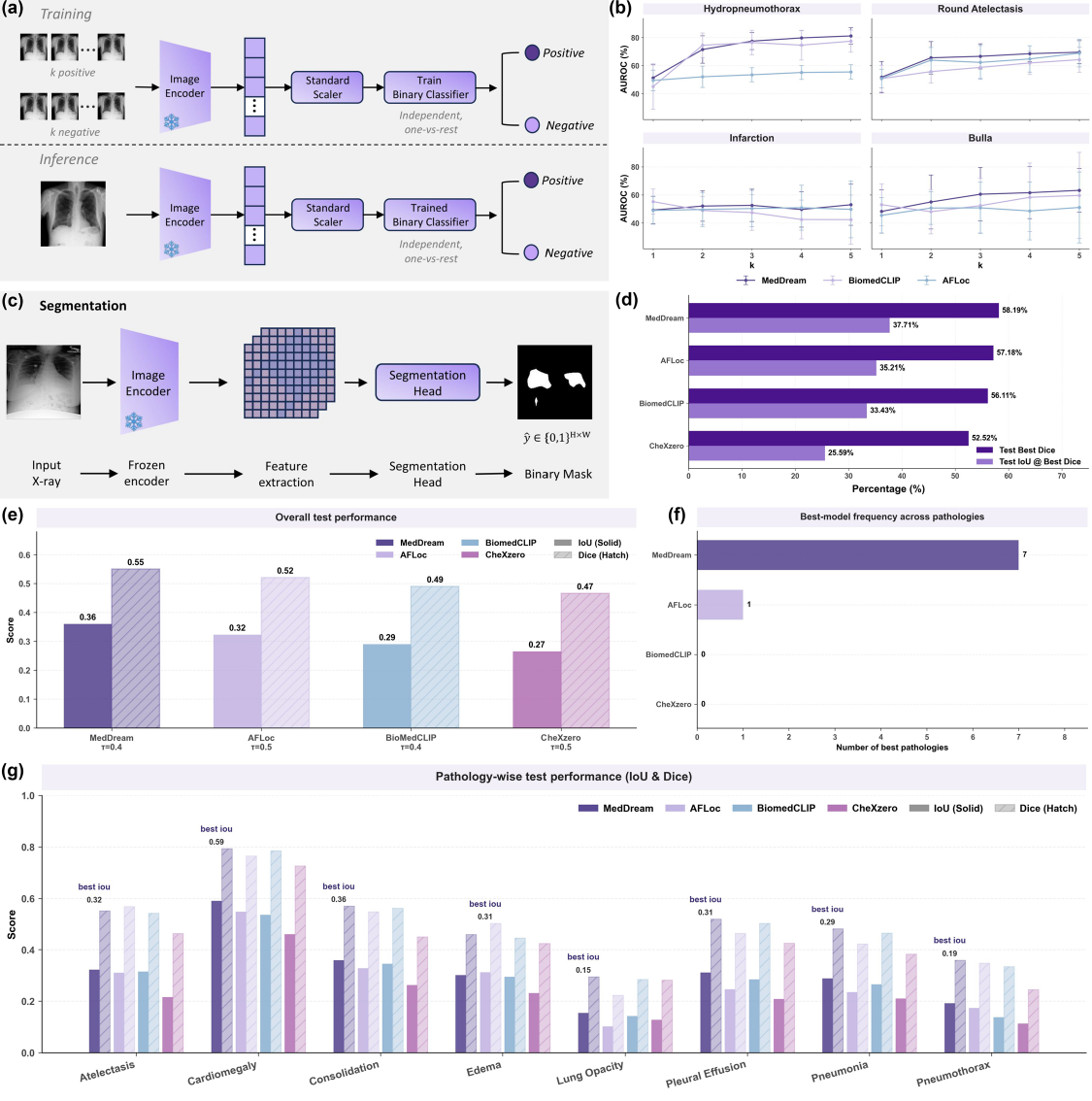}
\caption{\textbf{Few-shot adaptation and lesion-level grounding.}
\textbf{a,} Frozen-encoder k-shot classification protocol. For each uncommon finding, independent binary classifiers were trained using $k$ positive and $k$ negative examples, with $k=1$--5.
\textbf{b,} Few-shot AUROC for hydropneumothorax, round atelectasis, infarction and bulla across $k=1$--5 labelled examples.
\textbf{c,} Frozen-encoder segmentation protocol. Image features were extracted by a fixed encoder and passed to a lightweight segmentation head to predict binary abnormality masks.
\textbf{d,} RSNA Pneumonia localization performance measured by test IoU and Dice score.
\textbf{e,} Overall MS-CXR localization performance measured by test IoU and Dice score.
\textbf{f,} Number of MS-CXR abnormality categories for which each model achieved the best test IoU.
\textbf{g,} Pathology-wise localization performance on MS-CXR across eight abnormality categories, measured by test IoU and Dice score.}
\label{fig:fewshot_localization}
\end{figure}

\paragraph{Lesion-level localization.}
Pathologic localization is a central requirement for clinical imaging AI, because radiographic findings need to be spatially verified before they can support diagnosis, triage, or treatment planning. We therefore tested whether the frozen MedDream representation retained spatially localizable pathological information by training only a lightweight segmentation head on extracted features (Fig.~\ref{fig:fewshot_localization}c). Two complementary benchmarks were used: RSNA Pneumonia, which provides a larger single-disease localization setting, and MS-CXR, which provides phrase-grounding annotations across eight thoracic abnormality categories. On RSNA Pneumonia, MedDream reached an IoU of 37.7\% (95\% CI, 35.7--39.7\%) and a Dice score of 58.2\% (95\% CI, 56.0--60.4\%), compared with 35.2\% and 54.1\% for AFLoc, respectively (adjusted $P$ values of 0.006 and 0.002 for IoU and Dice, respectively; Fig.~\ref{fig:fewshot_localization}d). On MS-CXR, MedDream reached an IoU of 36.0\% (95\% CI, 33.7--38.3\%) and a Dice score of 55.0\% (95\% CI, 52.4--57.6\%), exceeding BiomedCLIP, the strongest comparator, which reached 32.8\% IoU and 51.4\% Dice (adjusted $P$ values of 0.005 and 0.003 for IoU and Dice, respectively; Fig.~\ref{fig:fewshot_localization}e). Pathology-wise analysis further showed that MedDream performed best for seven of eight MS-CXR abnormalities, including atelectasis, cardiomegaly, consolidation, lung opacity, pleural effusion, pneumonia, and pneumothorax (Fig.~\ref{fig:fewshot_localization}f,g). 

\paragraph{MedDream-supported severity review.}
We next asked whether this severity information was useful at the point of interpretation.  In a two-round reader study, three radiology residents graded 100 held-out frontal chest radiographs spanning four severity categories, first unaided and then, after a washout period, with access to the MedDream-estimated severity category, its confidence score and severity-matched synthetic reference examples (Extended Data Fig.~\ref{fig:extended_fairness_rebalancing2}a). Independent radiologist consensus served as the reference for agreement analysis. Concordance increased across all three readers during MedDream-supported review. Exact agreement increased from 34\% to 36\%, from 73\% to 85\% and from 62\% to 68\%, while mean absolute ordinal error decreased from 1.13 to 0.87, from 0.30 to 0.16 and from 0.44 to 0.35, respectively. Quadratic-weighted kappa increased from 0.243 to 0.482, from 0.868 to 0.935 and from 0.799 to 0.852 (Extended Data Fig.~\ref{fig:extended_fairness_rebalancing2}b). Supported reassessment moved more ratings closer to, than farther from, the radiologist reference for Reader~1 (23 versus 0; $P<0.0001$) and Reader~2 (17 versus 3; $P=0.0026$), with the same directional pattern for Reader~3 (18 versus 10; $P=0.1849$; McNemar test; Extended Data Fig.~\ref{fig:extended_fairness_rebalancing2}c). Representative cases showed complete correction, partial improvement and occasional movement away from the reference (Extended Data Fig.~\ref{fig:extended_fairness_rebalancing2}d). These findings show that MedDream-supported review was associated with higher resident agreement with radiologist-assessed severity.

\subsection*{Report-conditioned generation produces clinically aligned radiographs.}

\paragraph{Distributional fidelity and pathology consistency of generated radiographs.}
The coupling hypothesis underlying MedDream predicts that the representation supporting frozen diagnostic transfer should also support report-conditioned generation. We therefore tested whether the same clinical representation space could support generation of visually realistic and clinically aligned chest radiographs. Generation was evaluated on an independent held-out MIMIC-CXR p19 test set of 3,500 frontal radiographs using report-derived prompts held identical across models. MedDream was compared with ChexGen, a chest-radiograph-specific generative foundation model, and MINIM, a generalist multi-organ medical image generator, each under its native inference protocol; for MedDream, this comprised the internal best-of-$K$ trajectory selection described in Methods. MedDream produced the closest match to the real-image distribution in both feature spaces, reaching an XRV-FID of 0.31 (95\% CI, 0.28--0.35) and a CLIP-FID of 1.27 (95\% CI, 1.14--1.41), compared with 1.34 and 2.31 for ChexGen and 5.69 and 4.32 for MINIM, respectively (Fig.~\ref{fig:generation_synthetic_evidence}a). This fidelity was not obtained by collapsing onto a small set of prototypical images: intra-prompt MS-SSIM was lower for MedDream than for ChexGen or MINIM ($0.420 \pm 0.110$ versus $0.513 \pm 0.074$ and $0.591 \pm 0.104$), indicating greater structural variation, while pathology-profile correlation was higher ($0.618 \pm 0.290$ versus $0.421 \pm 0.381$ and $0.416 \pm 0.274$), indicating stronger preservation of prompt-specific diagnostic signal. Per-case differences were significant after Benjamini--Hochberg adjustment (all adjusted $P<0.001$).

\paragraph{Expert review and feature-space analysis of clinical grounding.}
Expert and feature-space analyses further supported the clinical grounding of the generated images. In a blinded review, four experts independently scored each generated image together with its corresponding report-derived prompt on a five-point scale from $-2$ to $+2$, where $-2$ indicated severe image artifacts or complete prompt mismatch and $+2$ indicated high image realism with full prompt-level agreement. For the two reviewers shown in the main figure, MedDream received higher scores than the competing models ($0.475 \pm 1.198$ and $0.700 \pm 1.363$ versus $-1.325 \pm 1.163$ and $-1.050 \pm 1.377$ for MINIM and $0.400 \pm 1.172$ and $0.550 \pm 1.413$ for ChexGen; Fig.~\ref{fig:generation_synthetic_evidence}c). The two additional reviewers similarly assigned MedDream the highest mean scores (Extended Data Fig.~\ref{fig:additional_generation_readers}). In XRV feature space, MedDream-generated images showed substantially greater overlap with real reference images than ChexGen-generated images (88.5\% versus 43.0\%; Fig.~\ref{fig:generation_synthetic_evidence}b). Generation-quality statistics, multiplicity-adjusted comparisons, expert-review results and feature-space overlap analyses are summarized in Supplementary Table~\ref{tab:generation_synthetic_utility_statistics}.

\subsection*{Synthetic diagnostic evidence improves downstream model development.}

\begin{figure}[!htbp]
\centering
\includegraphics[width=0.95\textwidth]{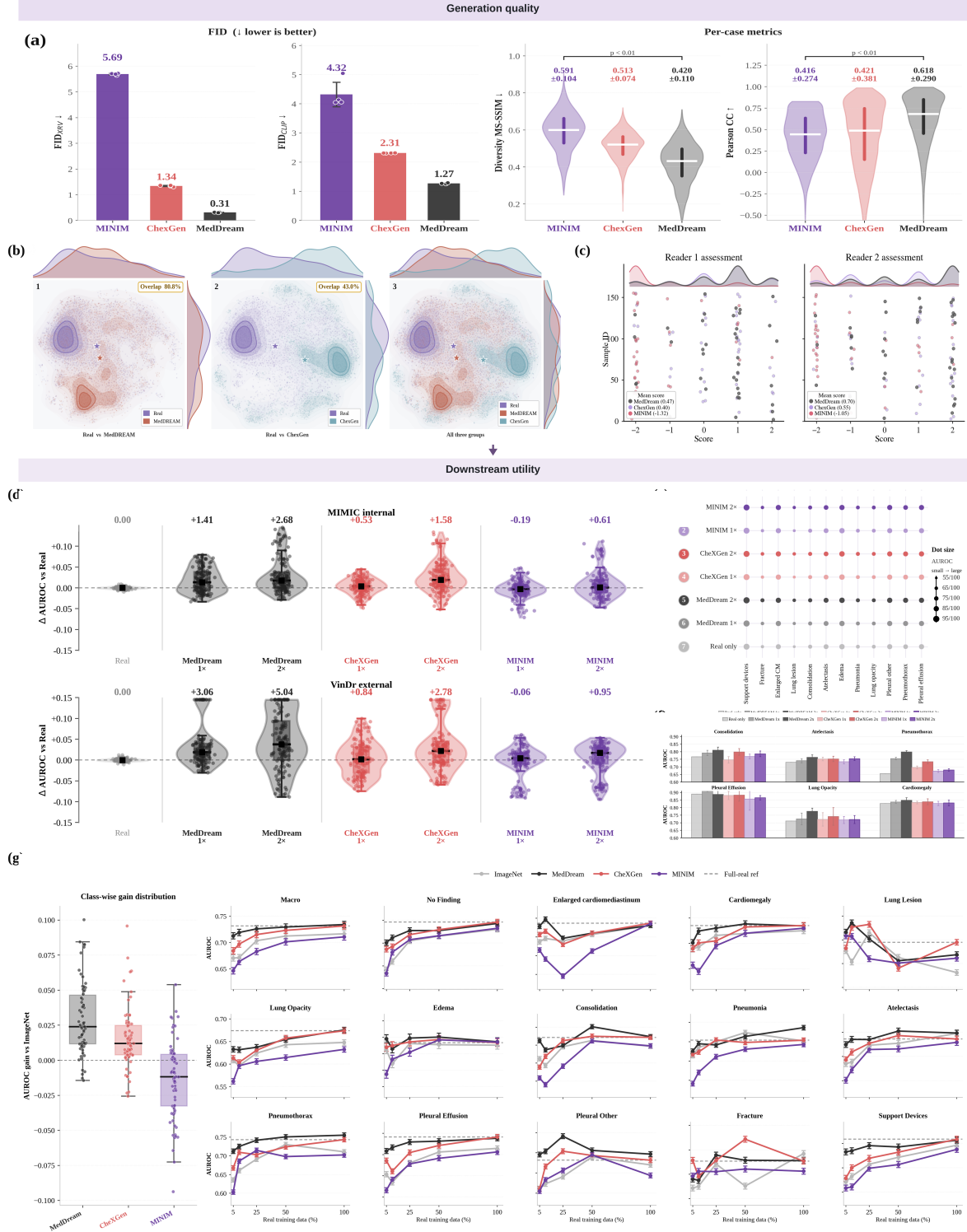}
\caption{\textbf{Report-conditioned generation and downstream synthetic-data experiments.}
\textbf{a,} Impression-conditioned generation quality evaluated by distributional fidelity and per-case metrics. FID was computed in XRV and CLIP feature spaces; pairwise MS-SSIM and Pearson correlation were computed at the case level.
\textbf{b}, Feature-space analysis of real and generated images using XRV DenseNet121 features projected by t-SNE. Feature-space overlap was computed from two-dimensional density estimates.
\textbf{c}, Blinded expert assessment of generated images using a five-point scale for image realism, acquisition plausibility and prompt-level consistency.
\textbf{d,} Change in macro-AUROC after 1$\times$ and 2$\times$ synthetic augmentation on internal MIMIC-CXR and cross-dataset VinDr-CXR classification.
\textbf{e,} Class-wise AUROC comparison across real-only training and synthetic augmentation with  MedDream, ChexGen and MINIM.
\textbf{f,} External VinDr-CXR AUROC by shared finding category after synthetic augmentation.
\textbf{g,} Synthetic-pretraining experiment across real-data budgets. Models were pretrained on synthetic images and fine-tuned using 5\%, 10\%, 25\%, 50\% or 100\% of the real training cohort.}
\label{fig:generation_synthetic_evidence}
\end{figure}

\paragraph{Improvement of internal and cross-dataset performance with synthetic diagnostic evidence.}
Clinical AI models are often trained on incomplete or institution-specific cohorts, limiting their exposure to heterogeneous disease appearances. We therefore tested whether report-conditioned synthetic images could provide reusable diagnostic evidence for real-image classification. Synthetic radiographs were generated from real training-study impressions and inherited the weak labels of their paired images, adding label-preserving visual variation rather than newly annotated cases (Methods). MedDream augmentation produced dose-dependent gains in both internal and cross-dataset evaluation (Fig.~\ref{fig:generation_synthetic_evidence}d). On the held-out MIMIC-CXR p19 test cohort, macro-AUROC increased from 72.3\% with real-only training (95\% CI, 71.1--73.5\%) to 73.7\% with $1\times$ augmentation (95\% CI, 72.5--74.9\%; adjusted $P=0.003$) and 75.0\% with $2\times$ augmentation (95\% CI, 73.8--76.2\%; adjusted $P<0.001$). Larger gains were observed on the fully external VinDr-CXR test set, where macro-AUROC increased from 76.4\% (95\% CI, 74.8--78.0\%) to 79.4\% with $1\times$ augmentation (95\% CI, 77.9--80.9\%; adjusted $P<0.001$) and 81.4\% with $2\times$ augmentation (95\% CI, 80.0--82.8\%; adjusted $P<0.001$). Under matched augmentation and optimization conditions, MINIM and ChexGen produced smaller and less consistent gains than MedDream on both test sets (Fig.~\ref{fig:generation_synthetic_evidence}e). On VinDr-CXR, gains were observed across all six shared findings, including increases of 14.3 percentage points for atelectasis (95\% CI, 9.7--18.9; adjusted $P<0.001$) and 4.3 percentage points for consolidation (95\% CI, 1.1--7.5; adjusted $P=0.009$) at $2\times$ augmentation (Fig.~\ref{fig:generation_synthetic_evidence}f). These results indicate that MedDream-generated images provided disease-relevant visual information that transferred to held-out real data.

\paragraph{Reduction of dependence on scarce real annotations through synthetic pretraining.}
Expert-labelled imaging cohorts remain a major bottleneck for clinical AI development, particularly for rare findings, underrepresented populations and institutions without large annotated archives. We therefore tested whether synthetic images could serve as a reusable diagnostic prior when real annotations were scarce. In a synthetic-pretrain-then-real-fine-tune setting, classifiers were first pretrained on model-specific synthetic images and then fine-tuned using 5\%, 10\%, 25\%, 50\% or 100\% of the real training cohort. Pretraining on MedDream-generated images improved data efficiency relative to ImageNet initialization, with a median AUROC gain of 2.4 percentage points across class--budget combinations (95\% CI, 1.7 to 3.5; adjusted $P=7.4\times10^{-12}$; Fig.~\ref{fig:generation_synthetic_evidence}g). The benefit was largest when real annotations were most limited. At the 5\% real-data budget, MedDream synthetic pretraining improved AUROC by 4.0 percentage points (95\% CI, 1.9 to 5.8; adjusted $P=1.2\times10^{-4}$), and at the 10\% budget by 4.9 percentage points (95\% CI, 3.1 to 6.4; adjusted $P=1.2\times10^{-4}$). The gain remained positive at the 25\% budget (1.4 percentage points; 95\% CI, 0.7 to 3.8; adjusted $P=4.0\times10^{-3}$). ChexGen produced smaller and more variable gains under the identical protocol, whereas MINIM produced predominantly negative gains. These results suggest that pretraining on MedDream-generated images transferred diagnostic structure to real-image classifiers and reduced dependence on large labelled cohorts.

\paragraph{Extension of synthetic evidence to quantitative prediction.}
We further tested whether synthetic diagnostic evidence extended beyond categorical disease classification to quantitative prediction. In a severity-regression setting using the RALO dataset, with 1,898 real training images and a fixed 475-image test set, MedDream augmentation produced monotonic improvements as synthetic-data volume increased across the three ratios tested. Mean absolute error (MAE) decreased from 0.808 with real-only training to 0.788, 0.774 and 0.751 with $1\times$, $2\times$ and $5\times$ MedDream augmentation, respectively; root mean squared error (RMSE) decreased from 1.008 to 0.981, 0.973 and 0.961; and the correlation between predicted and reference severity scores increased from 0.714 to 0.727, 0.735 and 0.746. At $5\times$ augmentation, MedDream reduced MAE to 0.751 (95\% CI, 0.717--0.785; adjusted $P<0.001$), reduced RMSE to 0.961 (95\% CI, 0.920--1.002; adjusted $P=0.004$) and increased the correlation to 0.746 (95\% CI, 0.713--0.779; adjusted $P=0.006$). MINIM and ChexGen produced smaller or less consistent improvements at matched augmentation scales. These findings indicate that MedDream-generated images provided dose-responsive signal for continuous prediction as well as categorical diagnosis, supporting their use in quantitative model development (Extended Data Fig.~\ref{fig:ralo_severity}).

\subsection*{Metadata-guided synthetic cohorts support subgroup stress testing and rebalancing.}

\begin{figure}[!htbp]
\centering
\includegraphics[width=\textwidth]{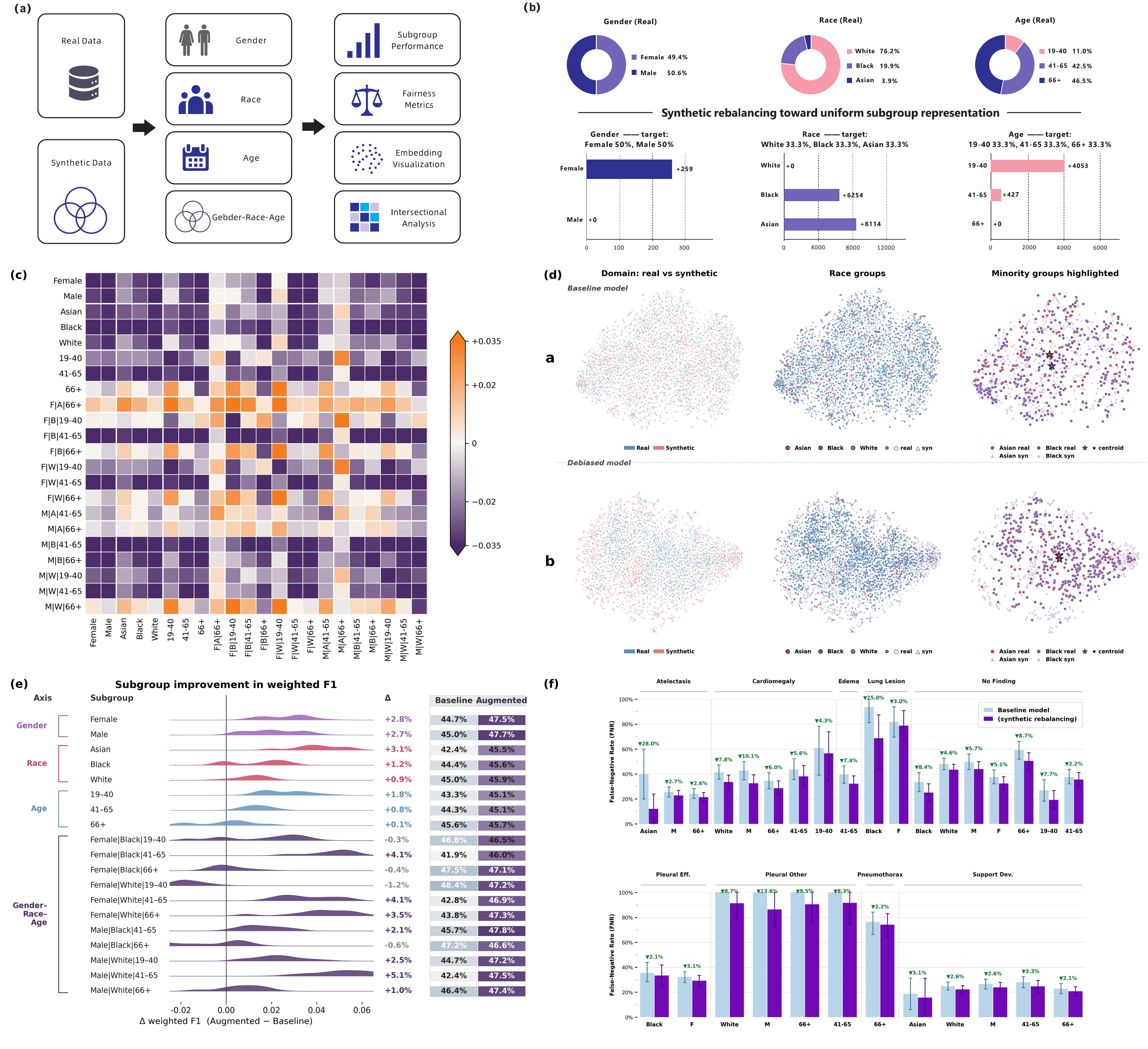}
\caption{\textbf{Metadata-guided synthetic cohorts for subgroup stress testing and targeted rebalancing.}
\textbf{a,} Subgroup evaluation framework. Real and synthetic cohorts were stratified by gender, race, age and intersectional demographic groups, followed by subgroup performance analysis, error-rate assessment, feature-space visualization and intersectional distribution analysis.
\textbf{b,} Synthetic rebalancing toward uniform subgroup representation across gender, race and age groups.
\textbf{c,} Wasserstein-distance matrix comparing real and synthetic subgroup disease-probability distributions across demographic subgroups, averaged over nine radiographic target labels. Lower values indicate closer subgroup distribution matching.
\textbf{d,} Descriptive t-SNE visualization of real and synthetic feature distributions before and after metadata-guided augmentation.
\textbf{e,} Change in weighted F1 after demographic synthetic augmentation across subgroup strata. Rows show gender, race, age and three-way intersectional groups.
\textbf{f,} False-negative-rate comparison before and after targeted synthetic augmentation across subgroup--pathology pairs.}
\label{fig:fairness_rebalancing}
\end{figure}

\paragraph{Preservation of subgroup-pathology structure in synthetic cohorts.}
A prerequisite for metadata-guided subgroup rebalancing is that added images preserve subgroup-pathology structure rather than merely increasing sample counts; otherwise, rebalancing may shift under-sampled groups towards majority-like disease patterns. We therefore assessed whether metadata-guided synthetic cohorts preserved pathology distributions across gender, race, age and their intersections (Fig.~\ref{fig:fairness_rebalancing}a). Rebalancing increased the representation of demographic strata that were sparse in the real training cohort, particularly across race and age groups (Fig.~\ref{fig:fairness_rebalancing}b). 

We next compared real and synthetic disease-probability distributions across 22 demographic groups using the mean Wasserstein distance over nine radiographic target labels. Matched real-synthetic subgroup pairs were substantially closer than unmatched pairs: the median Wasserstein distance was 0.051 (95\% CI, 0.047--0.062) for matched pairs and 0.141 for unmatched pairs (adjusted $P<0.001$). Mean distances showed the same pattern (0.057 versus 0.156; adjusted $P<0.001$), indicating that synthetic cohorts more closely resembled their corresponding real subgroups than other demographic groups (Fig.~\ref{fig:fairness_rebalancing}c). Higher distances were concentrated in very small intersectional strata, for which empirical disease distributions were less stable and were interpreted cautiously in subsequent analyses. 

Feature-space analyses provided an independent check for artefactual clustering or collapse towards majority groups. In two-dimensional t-SNE projections of penultimate-layer DenseNet-121 features, synthetic samples from under-sampled strata overlapped their corresponding real subgroups, and matched real-synthetic centroid distances were smaller than cross-group distances (Fig.~\ref{fig:fairness_rebalancing}d); these projections and centroid comparisons were descriptive and were not used for statistical testing. Nearest-neighbour and duplicate-screening analyses did not identify evidence of direct memorization of real training images. Together, these findings support the preservation of subgroup-specific disease structure in metadata-guided synthetic cohorts and establish the quality-control basis for the targeted rebalancing experiments below. Full Wasserstein-distance, feature-space and memorization analyses are reported in Supplementary Table~\ref{tab:subgroup_rebalancing_fairness_statistics}.

\paragraph{Reduction of subgroup missed diagnoses through metadata-guided rebalancing.}

We next tested whether metadata-guided synthetic rebalancing reduced subgroup-level diagnostic errors on held-out real images. We compared parity-targeted augmentation with real-only training and unguided synthetic augmentation, evaluating weighted F1 across demographic subgroups with at least 30 test cases (Fig.~\ref{fig:fairness_rebalancing}e,f). At $1\times$ targeted augmentation, the largest gains occurred in the sparsest race strata: weighted F1 increased by 3.1 percentage points in Asian patients (95\% CI, +1.0 to +5.2; adjusted $P=0.005$) and by 1.2 percentage points in Black patients (95\% CI, +0.1 to +2.3; adjusted $P=0.041$), whereas the change in White patients was not significant (+0.9 percentage points; adjusted $P=0.124$). Effects on sex and age were more heterogeneous: both sex strata improved similarly, while only patients aged 19--40 showed a significant gain (+1.8 percentage points; adjusted $P=0.028$). By contrast, unguided augmentation reduced F1 by 2.3 percentage points in Asian patients (adjusted $P=0.044$) but increased it by 3.7 percentage points in White patients (adjusted $P<0.001$). These results indicate that subgroup benefit depended on metadata-guided sample selection rather than synthetic-data volume alone, with effects varying across demographic axes.

False-negative-rate analysis at subgroup-pathology level identified reductions in some pairs and no changes or deterioration in others. Analyses were restricted to pairs with at least ten positive test cases, a threshold set before results were examined. For cardiomegaly, false-negative rates decreased from 0.427 to 0.326 in male patients (95\% CI, 0.354--0.500 to 0.258--0.393; adjusted $P<0.001$) and from 0.413 to 0.336 in White patients (95\% CI, 0.358--0.472 to 0.280--0.391; adjusted $P=0.005$). For atelectasis in Asian patients, the rate decreased from 0.400 to 0.120 (adjusted $P<0.001$), although this estimate rests on a small number of positive cases and a correspondingly wide interval (95\% CI, 0.200--0.600 to 0.000--0.240) and was therefore interpreted cautiously. Together with the subgroup-structure analyses above, these findings indicate that metadata-guided synthetic rebalancing reduced selected subgroup-level diagnostic errors and that its effects required evaluation at the subgroup-pathology level. All prespecified comparisons, including non-improving and worsening cells, together with weighted-F1 changes, false-negative-rate changes and multiplicity-adjusted statistics, are reported in Supplementary Table~\ref{tab:subgroup_rebalancing_fairness_statistics}.

\paragraph{Expert review showed fewer abnormal-finding omissions after rebalancing.}
To assess radiographically visible omissions beyond label-derived errors, we conducted a blinded expert review of 80 held-out cases enriched for subgroup-sensitive findings (atelectasis, cardiomegaly, ``no finding'' and four additional abnormal findings). Each reviewer first determined whether the prespecified target finding was present on the radiograph and then assessed anonymized outputs from the baseline and rebalanced models. The primary endpoint was abnormal-finding omission, defined as a reviewer-confirmed abnormal target finding reported as absent by the model; cases for which ``no finding'' was the prespecified target were excluded from this analysis. Reviewers~1, 2 and 3 confirmed 35, 36 and 34 abnormal-positive cases, respectively. Enrichment for findings on which the baseline model was prone to error means that the rates reported below exceed the label-derived false-negative rates observed on the full held-out test cohort and are not estimates of population-level missed-diagnosis frequency.


Within each reviewer's confirmed-positive set, the rebalanced model showed lower missed-diagnosis rates than the baseline model: 14.3\% versus 57.1\% (5/35 versus 20/35), 25.0\% versus 69.4\% (9/36 versus 25/36), and 14.7\% versus 64.7\% (5/34 versus 22/34) for Reviewers 1--3, respectively, corresponding to absolute reductions of 42.9, 44.4 and 50.0 percentage points (all adjusted $P\leq0.0003$, exact McNemar tests; Extended Data Fig.~\ref{fig:extended_reader_study}b). Paired case-level analysis identified 16, 16 and 17 corrected misses for Reviewers 1--3, respectively, with only one newly introduced miss across all three reviewers (Extended Data Fig.~\ref{fig:extended_reader_study}c--e). Descriptive subgroup analyses showed lower missed-diagnosis rates across gender and race strata, among patients aged 41 years or older, and for both atelectasis and cardiomegaly, whereas rates were unchanged in the 19--40-year group (Extended Data Fig.~\ref{fig:extended_reader_study}f). Read together with the label-derived false-negative analysis above, these findings indicate that rebalancing reduced radiographically confirmable omissions in cases selected to probe subgroup-sensitive failure modes.



\subsection*{Joint optimization couples diagnostic and generative learning.}

\begin{figure}[!htbp]
    \centering
    \includegraphics[width=\textwidth]{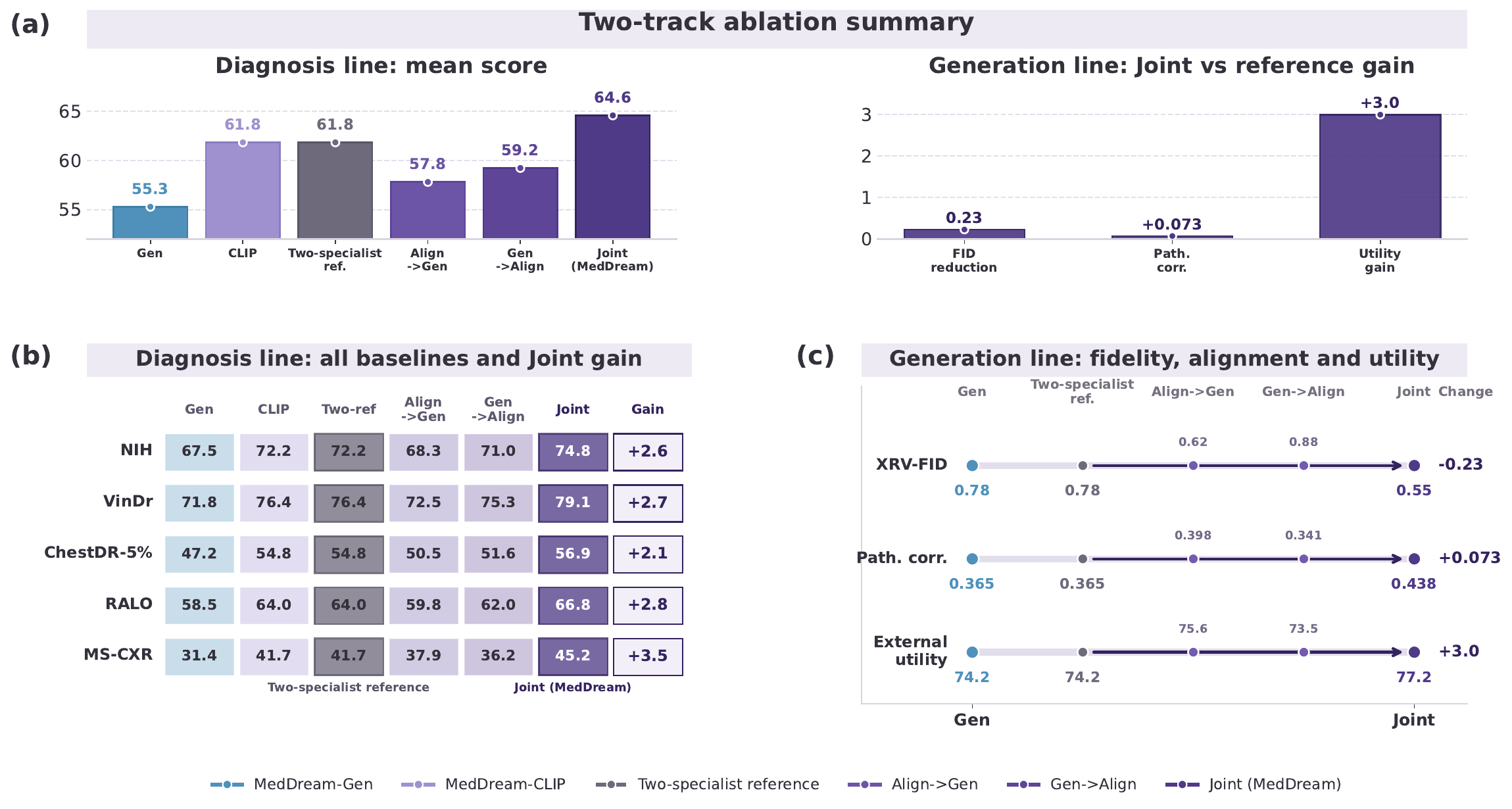}
    \caption{\textbf{Ablation analysis of the diagnostic and generative objectives.}
    MedDream-CLIP uses only CLIP-style image--text alignment, MedDream-Gen uses only masked latent generation without contrastive learning, and Full MedDream jointly optimizes both objectives through a shared visual pathway.
    \textbf{a,} Summary of diagnostic performance across single-objective, sequential and joint optimization strategies, together with generation-related gains of Full MedDream over MedDream-Gen. The
diagnostic summary score is the unweighted arithmetic mean of the
primary metric from each benchmark (AUROC for NIH, VinDr and
ChestDR-5\%; macro AUROC for RALO; Dice for MS-CXR).
    \textbf{b}, Diagnostic performance across chest radiography benchmarks.
    \textbf{c},  XRV-FID, pathology-profile correlation and external synthetic-data utility across generation-capable strategies.}
    \label{fig:two_track_ablation}
\end{figure}

Having established diagnostic transfer, spatially resolved representations, clinically grounded generation and downstream utility, we next asked whether these capabilities depended specifically on end-to-end joint optimization rather than on combining separately optimized models. We therefore compared single-objective, sequential and joint optimization strategies (Fig.~\ref{fig:two_track_ablation}). Joint optimization achieved the strongest performance across all five diagnostic and localization benchmarks, exceeding MedDream-CLIP by 2.1--3.5 percentage points. Compared with MedDream-Gen, joint optimization reduced XRV-FID from 0.78 to 0.55, increased pathology-profile correlation from 0.365 to 0.438 and improved external utility from 74.2 to 77.2. A post hoc two-specialist reference combining MedDream-CLIP and MedDream-Gen remained below the jointly optimized model across all evaluated endpoints. Sequential optimization also produced order-dependent trade-offs: Align$\rightarrow$Gen weakened diagnostic transfer, whereas Gen$\rightarrow$Align reduced generation fidelity and downstream utility. Together, these results show that neither separate specialists nor sequential optimization reproduced the balance achieved through end-to-end joint training, supporting joint optimization as the mechanism coupling diagnostic representation learning with clinically grounded generation in MedDream.

\clearpage
\section*{Discussion}
Clinical imaging datasets capture only a fraction of the visual evidence needed to develop and evaluate medical AI, particularly for rare disease presentations, newly defined diagnostic endpoints, and sparsely represented patient groups. Here, we show that transferable diagnostic representations and clinically grounded generation can be jointly learned through a shared radiographic state within a single radiographic world model. MedDream jointly optimizes image--text alignment and masked latent generation through a shared visual pathway, supporting label-efficient diagnostic readout and report-conditioned evidence generation. By coupling interpretation and simulation of radiographic observations, MedDream uses available clinical evidence while constructing plausible imaging evidence for settings in which comparable real-world observations are scarce.

Diagnostic and generative learning impose different masking requirements during pretraining while encouraging the shared representation to retain complementary clinical information. Diagnostic learning requires sufficient visible context to preserve disease-discriminative features, whereas generative learning benefits from stronger corruption that forces recovery of missing anatomy and pathology. These differing requirements motivate the progressive masking strategy. Prior hybrid designs have generally addressed this tension by separating diagnostic encoders from generative modules or by staging representation learning before generation~\cite{li2023blip,chu2025usp,li2026unified,zhang2026unix}, limiting the extent to which generative learning can shape the shared radiographic state. Our controlled optimization-strategy comparison indicates that this separation is both unnecessary and costly. Under matched data, architecture, and optimization budgets, joint optimization exceeded the alignment-only specialist across all five diagnostic and localization benchmarks by 2.1--3.5 percentage points, while also improving on the generation-only specialist in distributional fidelity, pathology-profile correlation, and downstream external utility. Sequential optimization did not recover this balance, instead producing order-dependent trade-offs in which each training order weakened the capability optimized first. A post hoc two-specialist reference also remained below the jointly optimized model across all evaluated endpoints, indicating that the advantage cannot be explained by simply combining two competent specialists. Progressive masking addresses this tension by exposing the encoder to a continuum of visible evidence, allowing semantic alignment and generative recovery to jointly shape the shared radiographic state.

Across held-out and pretraining-excluded evaluations, the shared radiographic representation retained complementary forms of clinical structure. Diagnostic transfer showed that MedDream preserved disease-related semantic information, while severity and localization analyses indicated retention of disease burden and spatial structure. The same representation also supported report-conditioned generation and alignment-guided trajectory selection. Compared with generation-only models, MedDream generated radiographs with stronger pathology consistency and closer agreement with real-image distributions, further supported by blinded expert review. Together, these findings indicate that joint training preserved discriminative clinical information while enabling clinically grounded generation within a shared radiographic state.

MedDream's value extends beyond visual plausibility. MedDream-generated images improved internal and cross-dataset performance, particularly when real annotations were limited, while synthetic pretraining suggested that they provided a reusable diagnostic prior rather than simply additional samples. Medical image generation should therefore be evaluated not only by realism, but also by whether generated observations preserve diagnostic information that transfers to real patient data. This utility depended on how synthetic evidence was selected and allocated. Compared with unguided augmentation, metadata-guided selection improved performance in prespecified subgroup--pathology settings and reduced selected abnormal-finding omissions on held-out real cases. These findings do not suggest that synthetic data can replace real clinical cohorts; rather, generated evidence should complement real data by addressing defined evidence gaps and supporting evaluation in sparsely or unevenly represented settings. Such evidence should remain tied to prespecified clinical questions and validated on independent real cases before informing deployment.

This study has several limitations. First, MedDream models examination-level radiographic state rather than longitudinal patient-state dynamics, and the present analysis was restricted to chest radiography. Extension to computed tomography and magnetic resonance imaging will require adaptation to volumetric structure and modality-specific acquisition processes, while longitudinal modeling will require temporally linked imaging and clinical data. Generated radiographs were not exhaustively assessed for all conditioned findings, anatomical localization, or unsupported abnormalities. The fixed-order severity reader study could not fully exclude familiarity or practice effects despite the washout period and independent re-randomization of case order. The supported condition was evaluated as a composite workflow combining model predictions, confidence scores, and synthetic reference images, and the contribution of individual components was not assessed separately. Confidence scores were not calibrated as clinical probabilities and should therefore be interpreted only as relative measures of model certainty. The literature-mined PMC-CXR component of the pretraining corpus retained an estimated 10.7\% residual rate of non-compliant pairs, and perceptual hashing may not detect all transformed variants of evaluation-set images; residual overlap between pretraining and evaluation data therefore cannot be fully excluded despite complementary metadata-based screening. Importantly, key diagnostic-transfer and downstream-utility findings were reproduced on datasets excluded entirely from pretraining, while both reader-study cohorts were independent of the public pretraining corpus. Demographic experiments relied on structured metadata and should therefore be interpreted as metadata-guided evidence selection rather than demographic-specific image generation. Although computational generalization was evaluated across independent datasets, the human reader studies were retrospective and conducted within a single health system, limiting conclusions regarding broader clinical utility across institutions and workflows. More broadly, radiographic world models should be evaluated not only by how accurately they interpret observed images, but also by whether the internal states they learn support reliable, verifiable, and clinically useful simulation of imaging evidence.


\clearpage
\section*{Methods}

\subsection*{Pretraining corpus}

We constructed a large-scale chest-radiography image-text resource comprising approximately 4.40 million pairs. The resource combined 3,475,100 literature-mined PMC-CXR image-caption pairs with approximately 0.93 million image-text pairs collected from public clinical chest-radiography repositories. Fig.~\ref{fig:data1} summarizes the construction, geographic distribution and source composition of this resource. From this resource, we derived a leakage-controlled subset for MedDream pretraining through image-quality filtering, evaluation-set exclusion, leakage screening, within-source deduplication and cross-source deduplication. After these procedures, the final MedDream pretraining subset comprised approximately 2.65 million chest-radiography image-text pairs, including approximately 1.80 million retained PMC-CXR pairs and approximately 0.85 million pairs from public clinical repositories.

\textbf{MIMIC-CXR.} MIMIC-CXR~\cite{johnson2019mimic} contains 377,110 chest radiographs linked to 227,827 free-text radiology reports collected at Beth Israel Deaconess Medical Center. We extracted the \textit{findings} or \textit{impression} section as paired text. Only the official training split was used for pretraining. MIMIC-CXR organizes studies into patient-group directories named p10 through p19 based on the leading digits of the de-identified subject identifier; each directory contains a mutually exclusive set of patients, so no patient appears in more than one group. The entire p19 patient subset, all CXR-LT evaluation images and all 1,162 MS-CXR images were excluded at the subject or study level, yielding 331,319 pretraining-eligible images.

\textbf{CheXpert Plus.} CheXpert Plus~\cite{chambon2024chexpert} contains 223,228 chest radiographs with radiology reports from Stanford Hospital. The \textit{findings} or \textit{impression} section was used as paired text.

\textbf{PadChest.} PadChest~\cite{bustos2020padchest} contains 160,868 chest radiographs with reports labelled across 174 findings from Hospital San Juan de Alicante. The \textit{findings} section was used as paired text.

\textbf{NIH ChestX-ray14.} NIH ChestX-ray14~\cite{wang2017chestx} contains 112,120 frontal chest radiographs with 14 disease labels. Because this dataset also served as a downstream evaluation benchmark, only the official training and validation partitions totalling 86,524 images were used for pretraining, and the official test set of 25,596 images was excluded. Label vectors from the training partition were converted into natural-language radiological descriptions using rule-based templates. Test-split labels were not used to construct any pretraining text.

\textbf{BRAX.} BRAX~\cite{reis2022brax} contains 40,967 chest radiographs with 14 disease labels. Label vectors were converted into template reports.

\textbf{IU X-Ray.} IU X-Ray~\cite{demner2016preparing} contains 7,470 chest radiographs with 3,955 reports from Indiana University. The \textit{findings} section was used as paired text. 

CheXpert Plus, PadChest, BRAX and IU X-Ray were not used in any downstream evaluation and were included in full after screening for cross-source near-duplicates against all evaluation cohorts by perceptual hashing.

\textbf{PMC-CXR.} To broaden the visual and linguistic coverage of pretraining, we constructed PMC-CXR from the PubMed Central Open Access Subset~\cite{pmc_open_access}, which contained 7,741,315 articles as of 28 February 2026. Figure-caption pairs were extracted using PubMed Parser~\cite{achakulvisut2020pubmed}. Candidate pairs were retrieved using chest-radiography terms, including ``chest X-ray'', ``CXR'' and ``PA view'', together with radiology-focused MeSH terms, yielding more than three million candidates. We excluded images smaller than $100 \times 100$~px and captions containing fewer than 10 words. Non-radiographic figures were further removed using a BiomedCLIP-based visual-content filter~\cite{zhang2023biomedclip} with a cosine-similarity threshold of 0.30. Near-duplicate removal was performed in two stages. Within PMC-CXR, we applied 64-bit perceptual hashing~\cite{zauner2010implementation} with a Hamming-distance threshold of 8. Potential overlap with evaluation datasets was then screened using perceptual-hash matching at a Hamming-distance threshold of 10, together with metadata-based exclusion of articles citing known evaluation-dataset DOIs or PhysioNet identifiers.

To validate the quality of the retained corpus, we conducted a manual audit on 2{,}000 randomly sampled image-caption pairs stratified by BiomedCLIP similarity quartile. Two annotators independently assessed each pair for three criteria: (i)~whether the image was a single-view chest radiograph rather than a non-CXR image, multi-panel composite or schematic (CXR precision); (ii)~whether the caption accurately described the corresponding image (caption-image alignment); and (iii)~absence of annotation overlays such as arrows or text labels. Disagreements were adjudicated by a board-certified radiologist. The audit estimated a CXR precision of 94.8\% (95\% CI, 93.7--95.7\%), a caption-image alignment rate of 92.6\% (95\% CI, 91.4--93.7\%), and residual multi-panel and overlay rates of 2.1\% and 3.7\%, respectively. Each of the 2,000 sampled images was queried against the full 1.80 million-image corpus by perceptual hashing at a Hamming-distance threshold of~8. No near-duplicate match was identified outside the image itself, giving a binomial 95\% upper confidence bound of $3/2{,}000 = 0.15\%$ on the per-image residual duplication rate. Perceptual hashing detects near-duplicate images but may not identify all transformed variants such as cropped, rotated, annotated or recompressed versions of evaluation-set images that fall below the Hamming-distance threshold. The metadata-based exclusion of articles citing known evaluation-dataset DOIs or PhysioNet identifiers provides a complementary safeguard but cannot guarantee complete removal of all derivative images. Images flagged during the audit were removed from the final corpus. Following these procedures, PMC-CXR comprised approximately 1.80 million image-caption pairs. PMC-CXR was used exclusively for pretraining; no PMC-CXR-derived captions or annotations were used as ground truth for downstream model selection or evaluation. Because the literature-mined component retains an estimated 10.7\% residual rate of non-compliant pairs, PMC-CXR is best characterized as a large-scale but imperfectly curated pretraining resource rather than a clean image-text corpus.

\begin{figure*}[h]
\centering
\includegraphics[width=\textwidth]{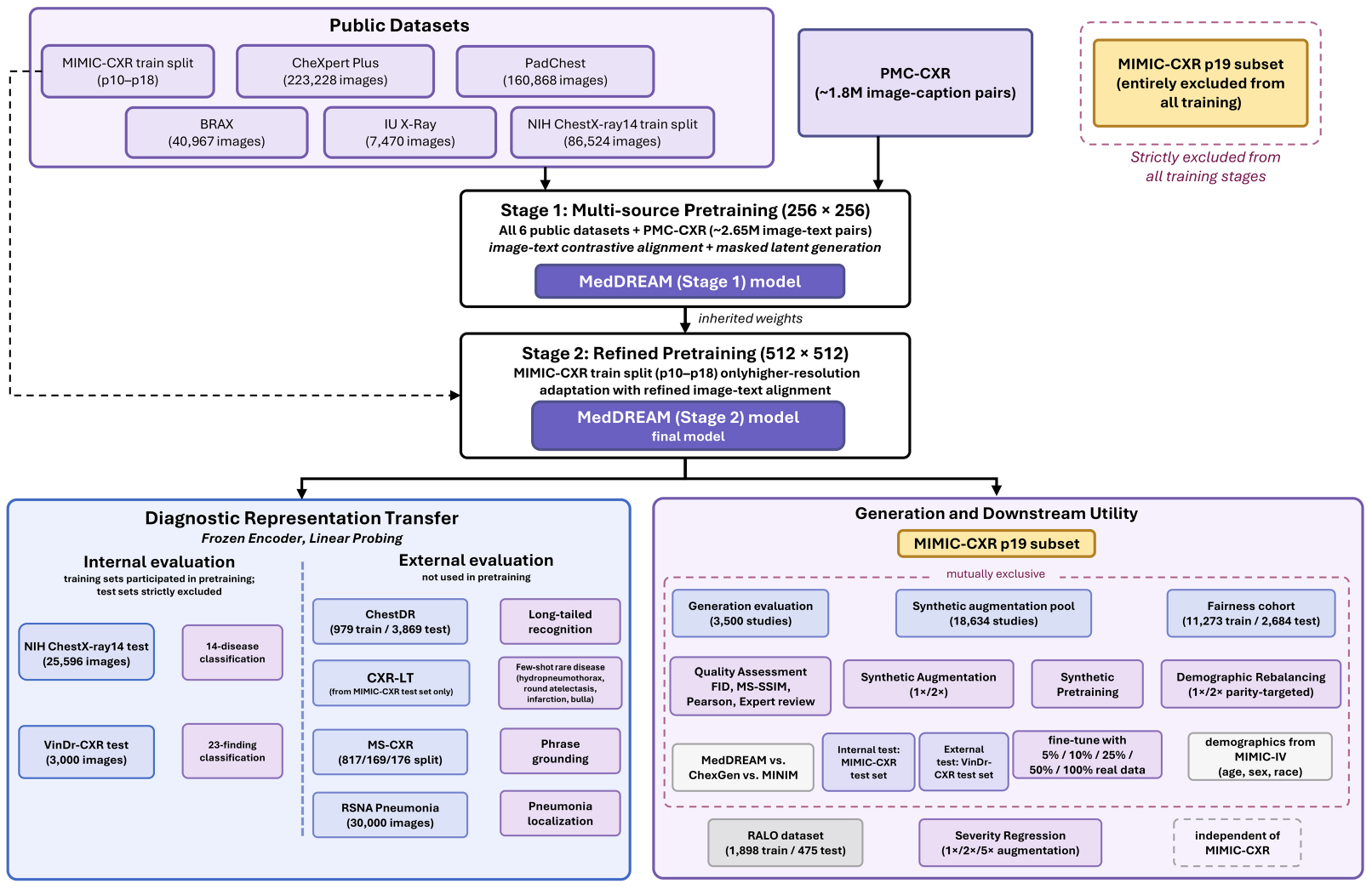}

\refstepcounter{extfigure}
\caption*{
\textbf{Extended Data Fig. \theextfigure | Schematic overview of MedDream training and evaluation.}
Stage 1 performs vision--language pretraining on six public chest-radiograph datasets and PMC-CXR, totaling approximately 2.65 million image--text pairs, using image--text contrastive alignment and masked latent generation. 
Stage 2 performs joint high-resolution refinement using only MIMIC-CXR patient groups p10--p18 while retaining both training objectives. 
The MIMIC-CXR p19 subset is excluded from MedDream pretraining and high-resolution refinement and is subsequently used for experiment-specific generation evaluation, synthetic-data development and metadata-guided rebalancing. 
Within each experiment, images used for downstream model development are separated from the corresponding real-image test cohort according to the partitioning protocol described in Methods. 
Diagnostic representation transfer is evaluated using frozen-encoder protocols across in-domain held-out and pretraining-excluded datasets. 
RALO is evaluated independently for radiographic severity prediction and synthetic-data augmentation.
}
\label{fig:data6}

\end{figure*}

\begin{figure*}[h]
\centering
\includegraphics[width=\textwidth]{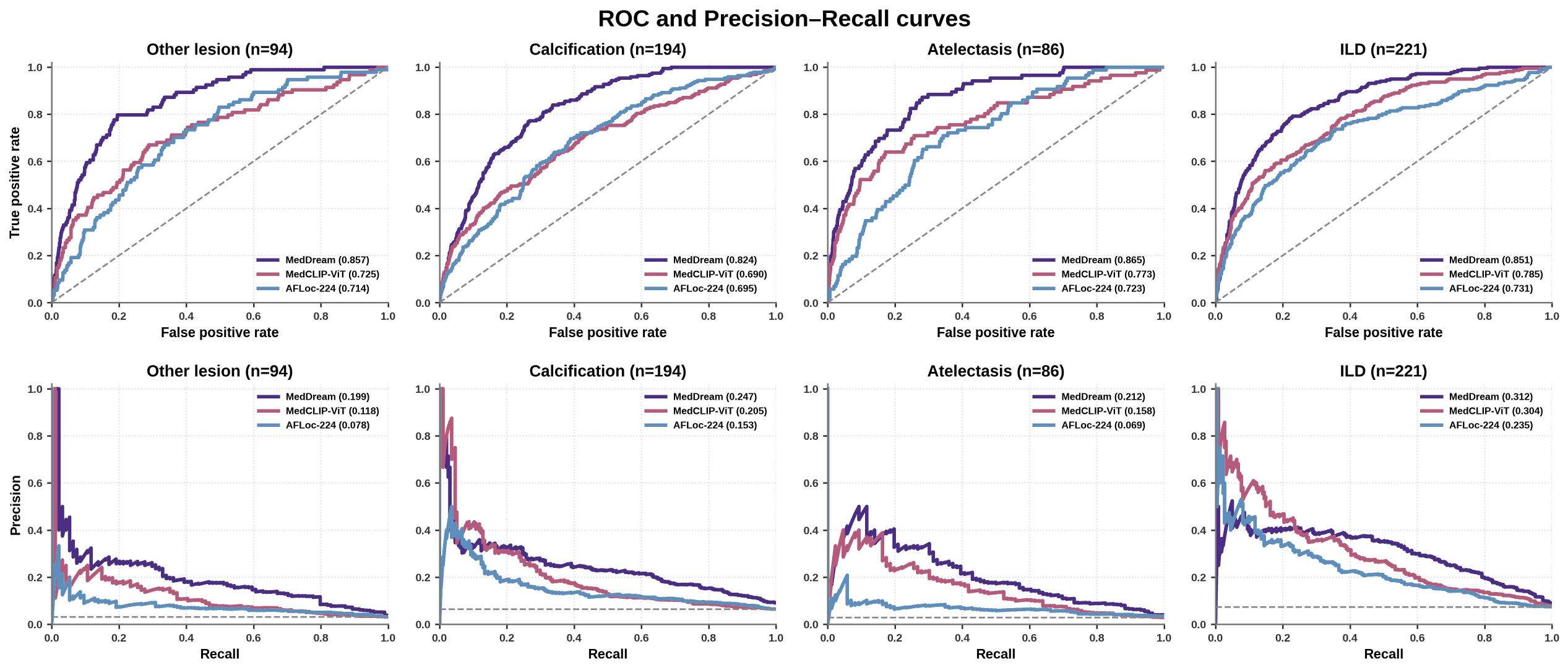}

\refstepcounter{extfigure}
\caption*{
\textbf{Extended Data Fig. \theextfigure | Extended evaluation of diagnostic adaptation.}
\textbf{a,} ROC curves for selected fine-grained VinDr-CXR labels.
\textbf{b,} Precision--recall curves for the same labels. Curves compare MedDream with MedCLIP and AFLoc using frozen-encoder linear probing.
}
\label{fig:extended_diagnostic_adaptation}

\end{figure*}

\begin{figure*}[h]
\centering
\includegraphics[width=\textwidth]{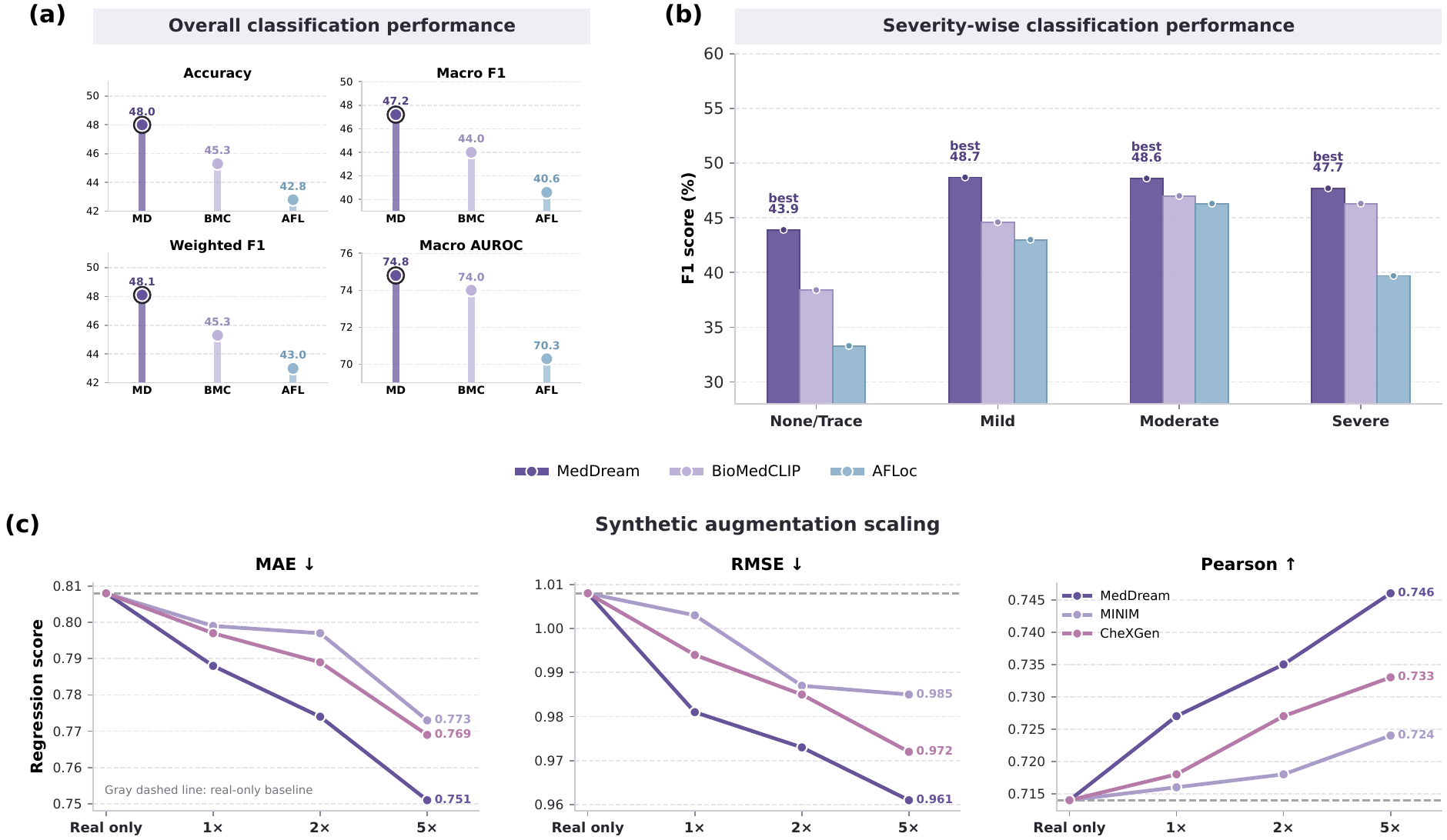}

\refstepcounter{extfigure}
\caption*{
\textbf{Extended Data Fig. \theextfigure | Radiographic severity classification and quantitative prediction on RALO.}
\textbf{a,} Overall severity-classification performance measured by accuracy, macro F1, weighted F1 and macro AUROC. Compared models include MedDream, BiomedCLIP and AFLoc.
\textbf{b,} Severity-wise F1 scores for none or trace, mild, moderate and severe categories.
\textbf{c,} Synthetic-augmentation scaling for quantitative prediction on RALO. Models were trained with real data alone or with 1$\times$, 2$\times$ and 5$\times$ synthetic augmentation, and evaluated using MAE, RMSE and Pearson correlation. Compared generative models include MedDream, MINIM and ChexGen.
}
\label{fig:ralo_severity}

\end{figure*}

\begin{figure*}[t]
    \centering
    \includegraphics[width=\textwidth]{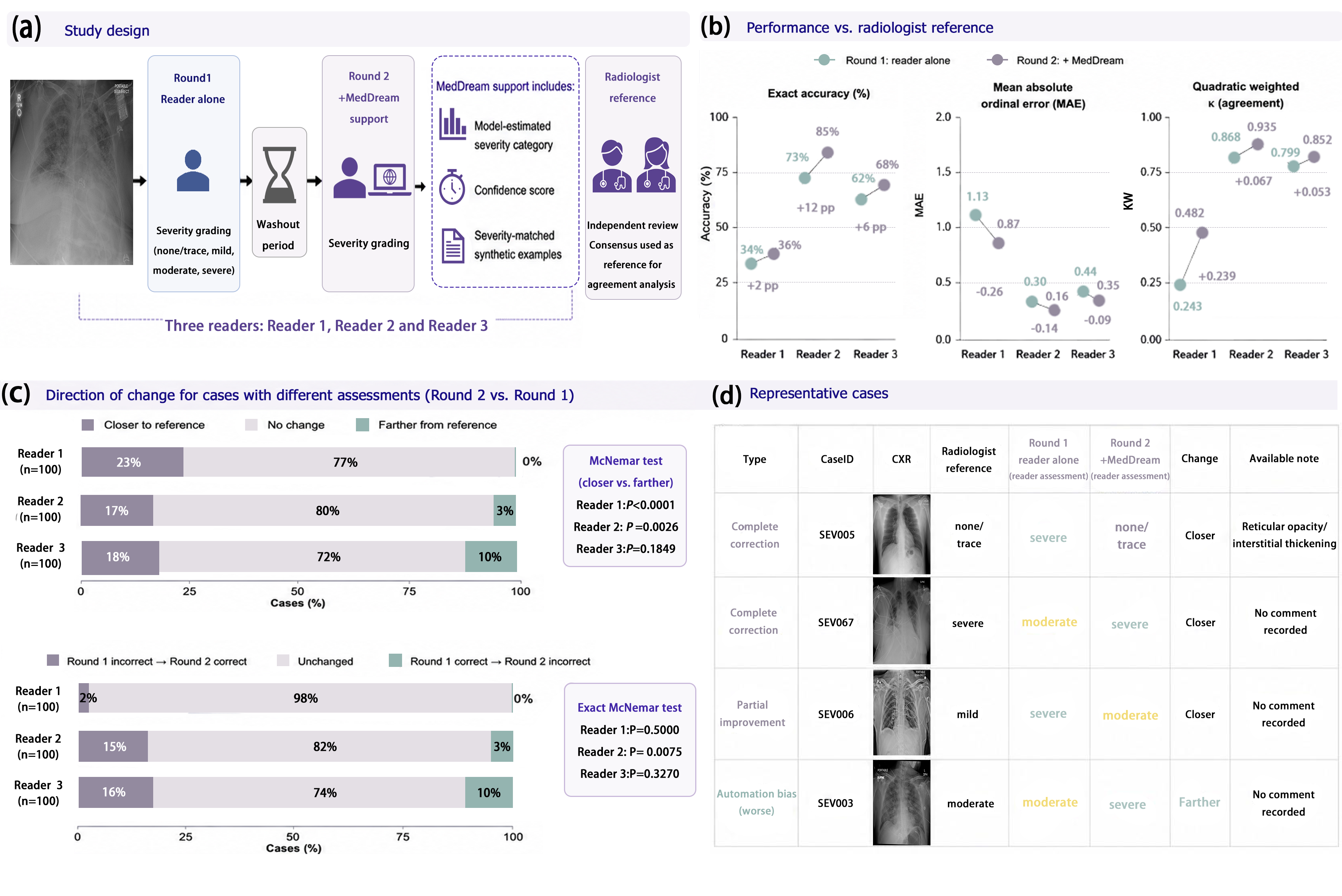}

    \refstepcounter{extfigure}
    \caption*{
    \textbf{Extended Data Fig. \theextfigure | MedDream-supported reader study for radiographic severity assessment.}
    \textbf{a,} Study design of the controlled reader study. Three readers first graded 100 held-out frontal chest radiographs without model support and, after a washout period, reassessed the same cases in randomized order with MedDream-supported evidence, including the model-estimated severity category, confidence score and severity-matched synthetic reference examples. Independent radiologist consensus was used as the reference for agreement analysis.
    \textbf{b,} Performance relative to the radiologist reference for each reader, including exact accuracy, mean absolute ordinal error (MAE) and quadratic-weighted kappa.
    \textbf{c,} Direction of assessment change from Round 1 to Round 2, showing the proportions of cases that moved closer to, remained unchanged relative to, or moved farther from the radiologist reference, together with exact-agreement error-correction analysis and paired McNemar tests.
    \textbf{d,} Representative cases illustrating complete correction, partial improvement and a shift farther from the radiologist reference after MedDream-supported review.
    }
    \label{fig:extended_fairness_rebalancing2}
\end{figure*}

\begin{figure*}[t]
    \centering
    \includegraphics[width=\textwidth]{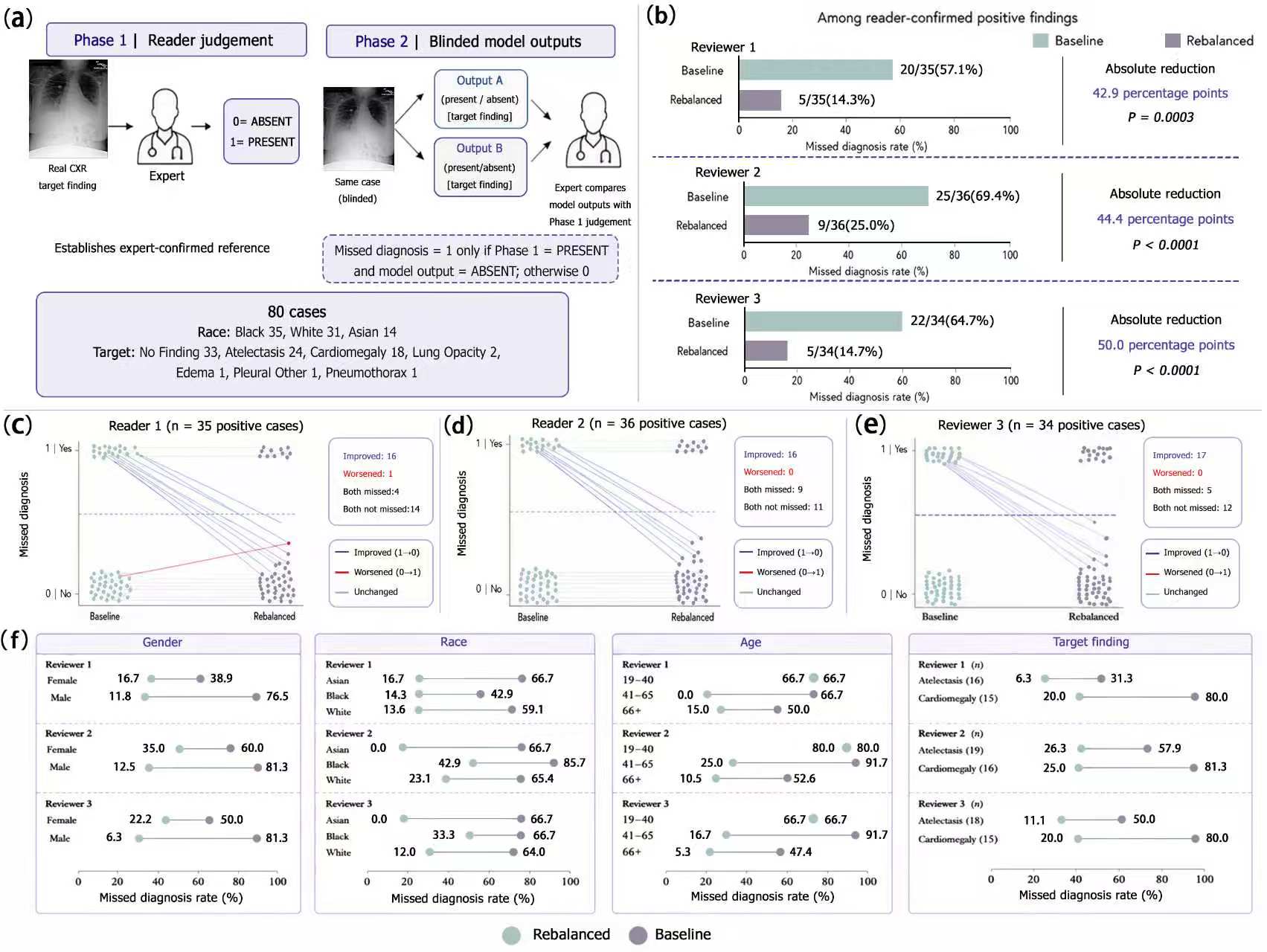}

    \refstepcounter{extfigure}
    \caption*{
    \textbf{Extended Data Fig. \theextfigure | Blinded expert adjudication of abnormal-finding omissions after model rebalancing.}
    \textbf{a,} Two-phase study design. Reviewers first determined whether the prespecified target finding was present on the chest radiograph and then assessed anonymized outputs from the baseline and rebalanced models. A false negative was recorded when a reviewer-confirmed target label was reported as absent by the model. For pathological findings this corresponds to a missed diagnosis; for the ``no finding'' category it represents a false-positive alert.
    \textbf{b,} Abnormal-finding omission rates among reviewer-confirmed positive cases for Reviewers~1--3. Absolute reductions are shown with \(P\) values from exact McNemar tests.
    \textbf{c--e,} Paired case-level transitions for Reviewer~1 (\textbf{c}), Reviewer~2 (\textbf{d}) and Reviewer~3 (\textbf{e}), including corrected, worsened and unchanged cases.
    \textbf{f,} Abnormal-finding omission rates across gender, race, age and target-finding subgroups.
    }
    \label{fig:extended_reader_study}
\end{figure*}

\begin{figure*}[t]
    \centering
    \includegraphics[width=\textwidth]{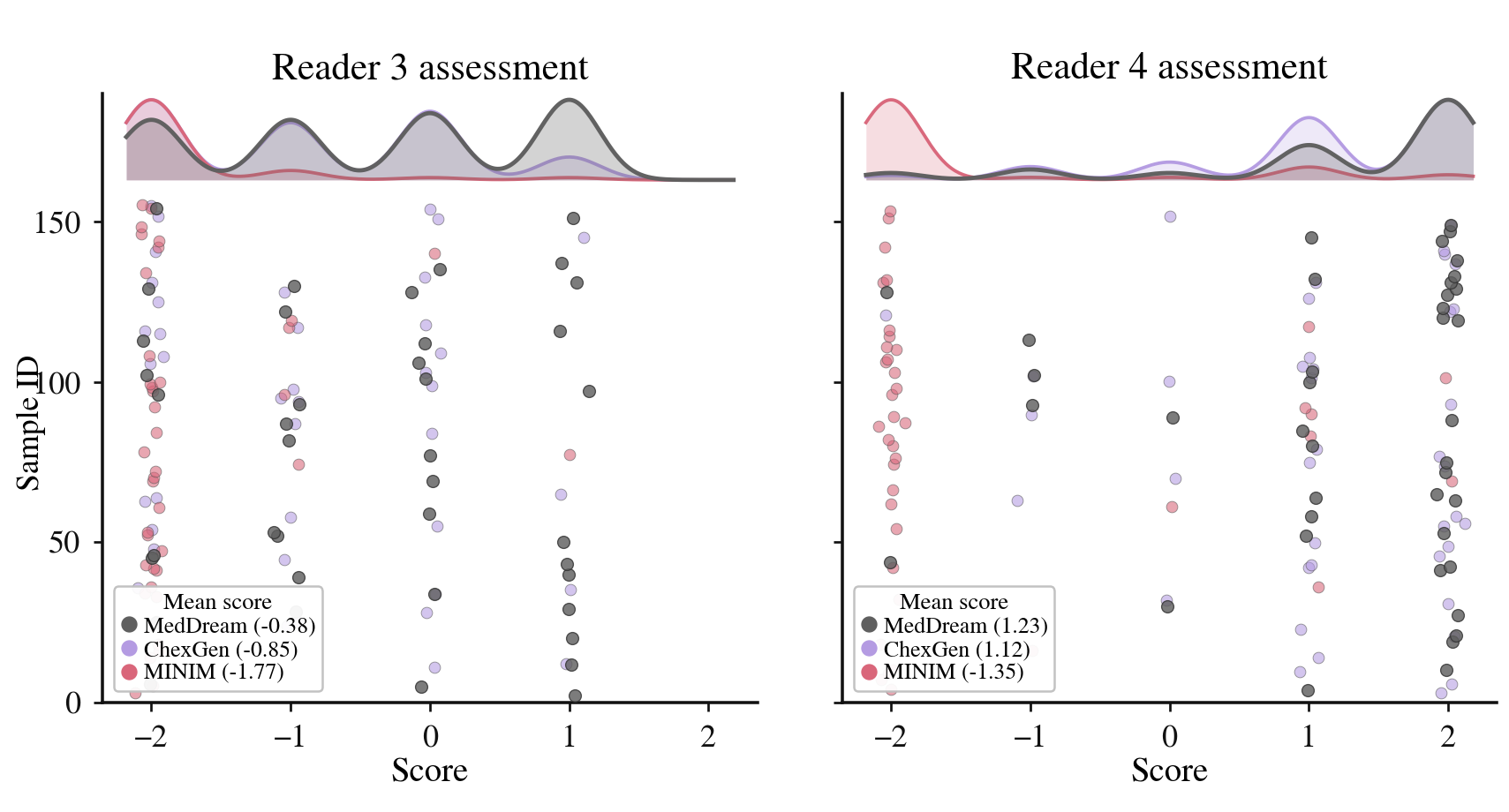}

    \refstepcounter{extfigure}
    \caption*{
    \textbf{Extended Data Fig. \theextfigure | Additional blinded expert assessment of report-conditioned radiographs.}
    Case-level scores assigned by two additional readers to images generated by MedDream, ChexGen and MINIM using a five-point scale from $-2$ to $+2$. Points represent individual images, upper density curves show the score distributions and legends report the mean score for each model. Higher scores indicate greater image realism and stronger agreement with the corresponding clinical report.
    }
    \label{fig:additional_generation_readers}
\end{figure*}

\textbf{Corpus assembly.} The pretraining-eligible public-repository
data, comprising approximately 0.85 million pairs, and PMC-CXR were
merged and deduplicated by perceptual hashing at a Hamming distance
threshold of 8. All images were resized to $256 \times 256$\,px and
normalized to zero mean and unit variance. Where patient or subject
identifiers were available, all exclusions were enforced at the
patient level. Otherwise, study- or image-level exclusion was applied
based on released benchmark manifests. A complete overview of all
data sources, their pretraining and evaluation partitions,
text-construction strategies and the exclusion protocol is provided
in Fig.~\ref{fig:data6}. Unless explicitly stated as post-hoc
analysis, all model selection, hyperparameter tuning, threshold
selection and early stopping used only training and validation data.

\subsection*{Datasets for diagnostic evaluation}
\textbf{NIH ChestX-ray14.} NIH ChestX-ray14~\cite{wang2017chestx}
was evaluated using the official split of 86,524 training and 25,596
test images across 14 pathological categories. The training partition
overlaps with the pretraining corpus, whereas the test partition was
excluded from pretraining. These 14 categories were treated as the
standard benchmark label space defined by NIH ChestX-ray14, without
applying a prevalence-based disease categorization.

\textbf{VinDr-CXR.} VinDr-CXR~\cite{nguyen2022vindr} contains
18,000 chest radiographs with 22 local and 6 global labels. This
dataset was not included in pretraining. The official split of 15,000
training and 3,000 test images was used for evaluation.
Multi-annotator labels were aggregated by majority vote. We evaluated
23 thoracic disease endpoints after excluding clavicle fracture,
edema and lung cyst because of insufficient positive cases. The
non-pathological labels ``no finding'' and ``other disease(s)'' were
retained only as dataset metadata and were not included as diagnostic
endpoints.

\textbf{RSNA Pneumonia.} RSNA
Pneumonia~\cite{rsna-pneumonia-detection-challenge} contains 30,000
frontal chest radiographs with lung-opacity and pneumonia labels.
This dataset was not included in pretraining.

\textbf{ChestDR.} ChestDR~\cite{Wang2023} contains 4,848 images
with 19 long-tailed thoracic disease labels, split into 979 training
and 3,869 test images. This dataset was not included in pretraining.
The long-tailed setting reflects the naturally imbalanced label
frequencies in ChestDR, while label scarcity was further evaluated
by training linear probes using 5\%, 10\%, 25\%, 50\% and 100\% of
the training set. The 5\% setting therefore contained approximately
49 total training images shared across all 19 labels.

\textbf{CXR-LT.}
We used the CXR-LT 2023 challenge data distributed as part of the CXR-LT PhysioNet v2.0.0 release~\cite{PhysioNet-cxr-lt-iccv-workshop-cvamd-2.0.0}, which defines a 26-class long-tailed disease label space derived from MIMIC-CXR. Four selected rare findings, namely hydropneumothorax, round atelectasis, infarction and bulla, were evaluated using an adapted two-way $k$-shot protocol based on the Ark+ evaluation framework~\cite{ma2025fully}. For each finding, $k$ positive and $k$ negative support examples were sampled, with $k$ ranging from 1 to 5. Evaluation was restricted to images from the official MIMIC-CXR test set, and all corresponding studies were excluded from MedDream pretraining.

\textbf{MS-CXR.} MS-CXR
v1.1.0~\cite{PhysioNet-ms-cxr-1.1.0} is a phrase-grounding benchmark
with 1,162 image-sentence pairs across eight cardiopulmonary
findings, divided into 817 training, 169 validation and 176 test
pairs. All MS-CXR images and their corresponding MIMIC-CXR studies
were excluded from the pretraining corpus at the study level. The
MS-CXR training split was used only for lightweight localization
heads. The MS-CXR validation split was used solely for
segmentation-threshold selection. No MS-CXR test annotations were
used for training, model selection, threshold selection or trajectory
selection.

\subsection*{Datasets for generation and downstream utility evaluation}

Impression-conditioned generation, synthetic classification utility and demographic subgroup-rebalancing experiments used data from the MIMIC-CXR p19
patient subset, which was excluded from pretraining in its entirety.
Within p19, non-overlapping study sets were defined at the study
level for each experiment, following the evaluation protocol of
ChexGen~\cite{ji2026generative}. Severity regression was evaluated
separately on RALO, an independent radiologist-scored lung-opacity
severity dataset not derived from MIMIC-CXR.

\textbf{Impression-conditioned generation.}
Impression-conditioned generation used 3,500 frontal p19 studies,
selected to be disjoint at the patient level from all p19 studies
used for synthetic augmentation, synthetic pretraining and downstream
classifier training. Prompts were derived from the impression or
conclusion sections of the corresponding held-out reports. No p19
image or report was used during MedDream pretraining or fine-tuning,
and reference images were never used as generation inputs.

\textbf{Synthetic augmentation and pretraining.} Synthetic augmentation and pretraining used 18,634 frontal p19
studies, disjoint at the patient level from the 3,500-study
generation set.
Synthetic images were generated exclusively from training-set
impressions. No test-set report was used to condition any synthetic
image. Internal evaluation was performed on a held-out real p19 test cohort
that was disjoint at the patient level from all p19 studies used for
prompt-conditioned synthetic generation, and external evaluation
was performed on the VinDr-CXR test set.

\textbf{Severity regression.} Severity regression used the RALO
dataset with 1,898 training and 475 test images scored by
radiologists for lung-opacity severity. Synthetic augmentation was
applied at 1$\times$, 2$\times$ and 5$\times$ scales.

\textbf{Demographic subgroup rebalancing.} Demographic fairness used a separate
p19 cohort of 11,273 training and 2,684 test images with
CheXpert-style labels, split at the patient level using
\texttt{subject\_id} before any synthetic data were generated.
Demographics linked from MIMIC-IV records were used only for
training-set targeted sampling and test-set subgroup evaluation.
Test-set distributions did not influence training sampling.
Parity-targeted augmentation at 1$\times$ and 2$\times$ scales was
compared with unguided baselines.

\subsection*{Datasets for clinical reader studies}

Two controlled reader studies were conducted using controlled-access
chest radiograph cohorts curated at the University of Chicago. These
cohorts were independent of all public datasets used for pretraining
and automated evaluation.

\textbf{Severity reader study.} A controlled-access cohort of
held-out frontal chest radiographs was curated at the University of
Chicago, spanning four radiographic severity levels: none or trace,
mild, moderate and severe. Radiology residents reviewed each case
first without model support and, after a washout period of at least
two weeks, reassessed the same cases in a freshly randomized order
with MedDream-supported evidence, including the model-estimated
severity category, a confidence score defined as the softmax
probability of the predicted class, and severity-matched synthetic
reference examples. The confidence score was not post-hoc calibrated
and should be interpreted as a measure of relative model certainty
across cases rather than as a calibrated probability estimate.
Independent radiologist consensus was used as the reference for
agreement analysis rather than as an absolute ground truth. All
readers completed the unsupported round before the supported round to
prevent model-provided severity estimates from anchoring unassisted
judgments. This fixed ordering means that second-round improvements
may partly reflect case familiarity or practice effects; a minimum
two-week washout, independent re-randomization of case order and
absence of inter-round feedback mitigate but do not eliminate this
concern. The supported condition combined all three evidence
components into a single intervention, testing the composite workflow
without isolating individual contributions. No image from this cohort
was included in model pretraining.


\textbf{Expert adjudication study.} A separate controlled-access
cohort of 80 chest radiographs was curated at the University of
Chicago and enriched for prespecified findings relevant to
subgroup-level evaluation, including atelectasis, cardiomegaly and no
finding. Expert reviewers first assessed the prespecified target
finding directly from the original radiograph and subsequently
reviewed anonymized outputs from the baseline and rebalanced
classifiers. Model outputs were compared with reviewer-confirmed
target-finding status at the subgroup--pathology level. For
pathological findings, a model output of absent when the reviewer
confirmed the target finding as present was defined as an
abnormal-finding omission; for the ``no finding'' category, the
corresponding error represents a false-positive alert. Because the
review set was enriched for prespecified findings, the resulting
omission rates are conditional on the enriched case mix and do not
estimate population-level error rates in unselected clinical
settings. No image from this cohort was included in model pretraining.

\subsection*{MedDream as a radiographic world model}

\paragraph{Overview.}
MedDream learns a shared latent radiographic state that supports medical visual representation learning and text-conditioned image generation within a single vision--language framework. Rather than coupling an image generator with an external vision--language reranker, MedDream learns both capabilities inside one model. It first maps medical images into spatially organized continuous latent tokens using a pretrained VAE encoder. A transformer encoder then infers a radiographic representation from the visible latent tokens, while a text-conditioned decoder performs masked autoregressive latent generation under the guidance of the learned semantic space. In parallel, an image--text alignment objective structures this representation according to clinical semantics. As a result, the same latent radiographic state supports both diagnostic readout and clinically conditioned image generation.

\paragraph{Continuous latent tokenization and visual encoding.}Given a medical image $x$, a pretrained VAE encoder $\mathcal{E}*{\mathrm{VAE}}$ first compresses the image into a spatially organized continuous latent feature map $F=\mathcal{E}*{\mathrm{VAE}}(x)\in\mathbb{R}^{C\times H'\times W'}$. The latent feature map is then flattened into $N=H'W'$ spatial latent tokens and projected to the model dimension as $X=\operatorname{Proj}(\operatorname{Flatten}(F))\in\mathbb{R}^{N\times d}$, where $d$ denotes the latent token dimension used by the transformer. During training, a mask set $\mathcal{M}\subset{1,\ldots,N}$ is sampled according to a masking ratio $r$, and we denote its complement as $\bar{\mathcal{M}}={1,\ldots,N}\setminus\mathcal{M}$. Only the visible latent tokens $X_{\bar{\mathcal{M}}}$, together with learnable buffer tokens $B_{\mathrm{buf}}$, are provided to the image encoder $E_{\theta}$, producing the visual representation $Z=E_{\theta}(X_{\bar{\mathcal{M}}},B_{\mathrm{buf}})$. The buffer tokens act as learnable global anchors that aggregate information from visible regions and provide a stable image-level representation for both clinical alignment and conditional generation. Clinical text is deliberately excluded from the image encoder. This design prevents the encoder from exploiting linguistic shortcuts and instead forces the model to learn anatomical structure, tissue texture, lesion morphology, and spatial context directly from the medical image. The resulting representation $Z$ therefore serves as the shared latent radiographic state used for both clinical readout and conditional generation.

\paragraph{Clinical image-text alignment.}To ensure that the visual representation carries clinical semantics rather than serving only as a reconstruction feature, MedDream introduces a medical image-text alignment branch. For each paired clinical text $y$, such as a radiology report, a finding description, or a disease prompt, the text encoder $T_{\phi}$ parameterized by $\phi$ produces a text representation $T=T_{\phi}(y)$. The image and text representations are then projected into a shared semantic space as $h_I=p_I(\operatorname{Pool}(Z))$ and $h_T=p_T(T)$, where $p_I$ and $p_T$ denote the image and text projection heads, respectively, and $\operatorname{Pool}(\cdot)$ denotes a pooling operation over visual tokens. For a mini-batch of $n$ paired image-text samples, the image-to-text contrastive loss is defined as
\begin{equation}
\mathcal{L}_{I\rightarrow T} = -\frac{1}{n} \sum_{i=1}^{n} \log \frac{\exp\left(\operatorname{sim}(h_I^i,h_T^i)/\tau\right)
}{\sum_{j=1}^{n} \exp\left(\operatorname{sim}(h_I^i,h_T^j)/\tau\right) }.
\end{equation}
where $\operatorname{sim}(\cdot,\cdot)$ denotes cosine similarity and $\tau$ is a temperature parameter. The text-to-image direction is computed symmetrically:
\begin{equation}
\mathcal{L}_{T\rightarrow I}
=
-\frac{1}{n}
\sum_{i=1}^{n}
\log
\frac{
\exp\left(\operatorname{sim}(h_T^i,h_I^i)/\tau\right)
}{
\sum_{j=1}^{n}
\exp\left(\operatorname{sim}(h_T^i,h_I^j)/\tau\right)
}.
\end{equation}

The final alignment objective is the average of the two directions, $\mathcal{L}_{\mathrm{align}}=(\mathcal{L}_{I\rightarrow T}+\mathcal{L}_{T\rightarrow I})/2$. Through this objective, MedDream structures the visual representation according to clinical concepts. Disease findings, anatomical locations, and imaging appearances are encouraged to establish stable correspondences with their paired textual descriptions, forming a clinically meaningful medical vision-language representation space.

\paragraph{Masked latent generation with diffusion reconstruction.} The generative pathway of MedDream is formulated as masked latent generation with a diffusion-based reconstruction head. Rather than directly regressing the values of masked latent tokens, the model recovers them by predicting the noise added to each masked continuous latent. For a masked latent token $X_i$, where $i\in\mathcal{M}$, we first construct a noisy latent token
\begin{equation}
X_{i,t} = \sqrt{\bar{\alpha}_t} X_i + \sqrt{1-\bar{\alpha}_t}\,\epsilon, \qquad \epsilon \sim \mathcal{N}(0,I),
\end{equation}
where $t\in\{1,\ldots,T_{\mathrm{diff}}\}$ denotes the diffusion timestep, $\bar{\alpha}_t=\prod_{s=1}^{t}\alpha_s$ is the cumulative noise schedule, and $\epsilon$ is Gaussian noise sampled from the standard normal distribution. Conditioned on the visible latent tokens $X_{\bar{\mathcal{M}}}$, the encoder representation $Z$, and the clinical text condition $c=T_g(y)$ produced by a generation text encoder $T_g$, the text-conditioned decoder uses a diffusion head $\epsilon_{\psi}$ parameterized by $\psi$ to predict the injected noise for each masked latent token. Here, $T_g$ denotes the text encoder used for generative conditioning, whereas the alignment text encoder $T_{\phi}$ is used to construct the contrastive image-text representation space. The corresponding diffusion reconstruction loss is defined as
\begin{equation}
\mathcal{L}_{\mathrm{diff}} = \mathbb{E}_{i\in\mathcal{M},\,t,\,\epsilon} \left[ \left\| \epsilon - \epsilon_{\psi} \left( X_{i,t}\mid t,\;X_{\bar{\mathcal{M}}},\;Z,\;c \right) \right\|_2^2 \right],
\end{equation}
where the expectation is taken over masked token indices $i\in\mathcal{M}$, diffusion timesteps $t$, and Gaussian noise $\epsilon$, and $\|\cdot\|_2^2$ denotes the squared Euclidean error.

This design can be understood as a two-level generation mechanism. The outer masked latent generation process determines which spatial latent tokens should be recovered at each step, while the inner diffusion head determines how to generate the continuous values of the selected masked tokens. Therefore, MedDream is not a conventional image-level diffusion model that directly maps noise to pixels or full latent maps, nor is it a standard autoregressive model that deterministically predicts token values. Instead, it integrates diffusion-based continuous latent reconstruction into a masked latent generation framework. More importantly, the generation process is constrained by a shared medical vision-language representation space. The diffusion reconstruction mechanism specifies how missing continuous latents are recovered, whereas the shared representation space determines what clinical semantics the recovered latents should express. In this way, image synthesis is not an isolated generation module, but is tightly coupled with the same representation space used for medical understanding.

The overall training objective jointly optimizes diffusion-based latent generation and clinical semantic alignment:
\begin{equation}
\mathcal{L}
=
\mathcal{L}_{\mathrm{diff}}
+
\lambda
\mathcal{L}_{\mathrm{align}},
\end{equation}
where $\lambda = 0.005$. Let $\hat{r} = |M|/N$ denote the realized
masking ratio after converting the sampled continuous masking rate to
an integer number of masked tokens via $|M| = \lceil N \cdot r
\rceil$. A single masking ratio was sampled for each mini-batch,
while the spatial masking order was sampled independently for each
image. The two objectives were gated by $\hat{r}$: the alignment
loss was computed only when at least 25\% of the visual tokens
remained visible ($\hat{r} \leq 0.75$), and the masked
autoregressive diffusion loss was computed only when at least 50\% of
the visual tokens were masked ($\hat{r} \geq 0.50$). The per-batch
objective was therefore
\begin{equation}
  \mathcal{L}(\hat{r})
  = \mathbf{1}[\hat{r} \geq 0.50]\,\mathcal{L}_{\mathrm{diff}}
  + \lambda\,\mathbf{1}[\hat{r} \leq 0.75]\,\mathcal{L}_{\mathrm{align}}.
\end{equation}
Fully visible inputs ($\hat{r} = 0$) contributed only to image-text
alignment, with the diffusion loss defined as zero without
normalization over an empty masked-token set. Fully masked inputs
($\hat{r} = 1$) contributed only to text-conditioned latent
generation through the generation text encoder $T_g$, with the
alignment loss omitted because no image tokens were available for
visual embedding. Both objectives were active in the overlapping
interval $0.50 \leq \hat{r} \leq 0.75$. The image encoder
$E_{\theta}$ is shared by the alignment and generation pathways and
receives gradients from both $\mathcal{L}_{\mathrm{align}}$ and
$\mathcal{L}_{\mathrm{diff}}$. The alignment text encoder
$T_{\phi}$ provides textual representations for the CLIP-style
image-text alignment objective, whereas the generation text encoder
$T_g$ provides the clinical text condition for masked latent
generation. The text-conditioned decoder and diffusion head are
optimized through the diffusion reconstruction objective. Through
this joint optimization, MedDream learns both the anatomical and
pathological structure of medical images, while also establishing
clinically meaningful correspondences between visual patterns and
textual concepts, yielding a shared radiographic state that supports both diagnostic readout and conditional generation.

\paragraph{Progressive masking curriculum.}Medical semantic alignment and image generation favor different masking regimes. A low masking ratio preserves sufficient visual context and is therefore beneficial for learning stable medical semantic representations. In contrast, a high masking ratio forces the model to recover missing latent regions from limited context, which is important for learning the conditional latent distribution required for generation. To reconcile these two objectives, MedDream adopts a masking warmup strategy. At epoch $e$, the masking ratio is sampled from a truncated Gaussian distribution, $r_e\sim\operatorname{TruncNormal}(\mu_e,\sigma^2;0,1)$, where the mean gradually increases as $\mu_e=\min(1,e/E_w)$. Here, $E_w$ denotes the number of warmup epochs. During early training, the model mainly operates under low masking ratios, allowing the encoder to establish a stable clinical image-text semantic space. As training progresses, the expected masking ratio increases, and the model is gradually exposed to more difficult high-mask recovery tasks. This transition encourages the model to move from semantic anchoring to generative recovery, allowing representation learning and generation learning to develop in the same continuous latent space.

\paragraph{Semantic trajectory selection.}
The shared medical vision-language space is reused during generation for semantic trajectory selection. Given a clinical prompt~$y$, MedDream initializes $K=8$ stochastic generation trajectories using distinct random seeds from 1 to~8. Each trajectory is advanced for $s_0=16$ early recovery steps out of $S=256$ total autoregressive iterations, producing a partial latent candidate~$X_k^{(s_0)}$. The partial candidates are not decoded into complete images for external reranking. Instead, each candidate is passed through the jointly trained image encoder, and its visual embedding is computed as $u_k=p_I!\left(\operatorname{Pool}!\left(E_{\theta}!\left(X_k^{(s_0)}\right)\right)\right)$. The clinical prompt is encoded by the alignment text branch as $v=p_T!\left(T_{\phi}(y)\right)$. Semantic consistency is then measured in the jointly trained image-text representation space:

\begin{equation}
  a_k = \operatorname{sim}(u_k, v),
  \qquad
  k^\ast = \operatorname*{arg\,max}_{k\in\{1,\ldots,K\}}a_k, 
\end{equation}
Only the selected trajectory~$X_{k^\ast}^{(s_0)}$ is continued for
the remaining $S - s_0 = 240$ recovery steps, after which the
completed latent representation is decoded by the VAE decoder.
This procedure is an internal best-of-$K$ candidate-selection
mechanism that differs from conventional external reranking in two
respects: it evaluates partial latent candidates rather than fully
rendered images, and it uses the same jointly trained representation
space that supports diagnostic transfer rather than an independently
trained classifier or vision-language model. Thus, semantic trajectory selection constitutes an intrinsic inference capability of the dual-foundation representation, rather than an external post-hoc reranking procedure. Only the selected trajectory is completed and decoded, whereas the remaining candidates are evaluated at an early partial-latent stage and are not rendered as full radiographs. Nevertheless, it
evaluates multiple stochastic candidates per prompt and incurs
additional inference cost. Under the default configuration the
procedure performs $K\,s_0 + (S - s_0) = 368$ autoregressive recovery
steps compared with~256 for single-trajectory generation, a 43.8\%
increase in recovery-step count, and additionally requires one
image-encoder forward pass for each of the eight partial candidates.
Because step count alone does not capture differences in per-step
computation, encoder evaluation or parallel execution, the
recovery-step count should be interpreted as an approximate measure
of relative computational cost rather than as a direct estimate of
wall-clock time.
All generation results reported in this study were obtained under
each model's native inference procedure. For MedDream, this includes
the partial-latent trajectory-selection procedure described above.
MINIM and ChexGen were each run with their published default
inference settings using a single seed per prompt.

This mechanism reflects the joint diagnostic-generative design of
MedDream. During training, image-text alignment organizes the visual
representation according to clinical findings, anatomical locations
and imaging appearances. During generation, the same representation is
reused to rank partially recovered latent trajectories according to
their semantic consistency with the clinical prompt. Rather than
relying on an external reranker, MedDream performs candidate selection
inside the jointly trained representation space, while retaining the
general best-of-$K$ principle of evaluating multiple stochastic
candidates.

\subsection*{General evaluation protocol}

All evaluations were conducted on held-out test data that were excluded from the corresponding pretraining, fine-tuning and synthetic-generation cohorts. Where a benchmark contributed data to pretraining, inclusion was restricted to its designated model-development partition; downstream probes were fitted using the benchmark training data and evaluated exclusively on the held-out test partition. No test-set image, report, label or metadata was used for pretraining, model selection, threshold selection or hyperparameter tuning. Patient-level separation was enforced for all datasets providing
patient or subject identifiers, including all MIMIC-CXR-derived
experiments. Study-level separation was used only when patient
identifiers were unavailable.

Unless otherwise specified, diagnostic representation transfer was evaluated using a frozen-encoder protocol. Pretrained image encoders were kept fixed, and only lightweight task-specific classifiers, regression heads or segmentation heads were trained on extracted features. This design was used to isolate the diagnostic information contained in the pretrained representation from the capacity of downstream task-specific models.

Classification performance was summarized using area under the receiver-operating-characteristic curve (AUROC), average precision, F1 score and Matthews correlation coefficient where applicable. Macro-level scores were computed by averaging metrics across evaluated disease labels. Severity classification was evaluated using accuracy, macro F1, weighted F1 and macro AUROC. Segmentation performance was assessed using intersection-over-union and Dice score. Quantitative prediction was evaluated using mean absolute error, root mean squared error and Pearson correlation coefficient.

For synthetic-data experiments, generated images were evaluated both as images and as training evidence. Image--level evaluation assessed distributional fidelity, intra-prompt diversity, pathology-profile consistency and expert-rated clinical plausibility. Downstream utility was assessed by training classifiers or regression models with real data alone or with matched synthetic augmentation, followed by evaluation on held-out real test sets. Subgroup analyses were performed only on held-out real images, with demographic groups and subgroup--pathology pairs filtered according to prespecified sample-size criteria.

\subsection*{Frozen-encoder diagnostic representation transfer}

Diagnostic representation transfer was evaluated using a frozen-encoder linear-probing protocol. For each model, the pretrained image encoder was fixed and used to extract image-level features from training, validation and test images. Features were standardized using statistics estimated from the training split, and the same transformation was applied to validation and test data. For multi-label classification tasks, disease-wise one-versus-rest logistic-regression classifiers were trained on frozen features with class-balanced weighting. For severity classification, a multinomial logistic-regression classifier was trained to predict ordinal radiographic severity groups. This protocol isolated the diagnostic information contained in pretrained image representations from the capacity of high-capacity downstream networks.

Common thoracic disease recognition was evaluated on NIH ChestX-ray14 using the official 86,524-image model-development partition and the held-out test partition of 25,596 images. The model-development partition contributed to MedDream pretraining and was subsequently used to fit disease-specific linear probes on frozen image features. The official test partition was excluded from pretraining and used only for final evaluation. NIH ChestX-ray14 was therefore treated as an in-domain held-out benchmark rather than a fully unseen-domain evaluation. Labels were derived from the \texttt{Finding Labels} field and covered 14 thoracic disease categories. The primary endpoint was mean AUROC across the 14 labels. Adaptation to a broader diagnostic label space was evaluated on VinDr-CXR using its official 15,000-image training partition and 3,000-image held-out test partition. Neither partition was included in MedDream pretraining; VinDr-CXR was therefore treated as a pretraining-excluded benchmark for assessing adaptation to an expanded downstream label space. The training partition was used to fit the downstream linear probes, and the test partition was reserved for final evaluation. VinDr-CXR includes both image-level findings and fine-grained local lesion labels, allowing us to assess adaptation to diagnostic endpoints defined for downstream use. Performance was reported using disease-wise AUROC and mean AUROC across the evaluated labels. Precision-recall curves were additionally generated for low-prevalence and fine-grained findings, for which average precision provides complementary information under class imbalance.

Label efficiency was assessed on ChestDR, a thoracic disease dataset with 19 disease labels and a long-tailed class distribution. Linear probes were trained using 5\%, 10\%, 25\%, 50\% and 100\% of the available training data. At each label fraction, frozen features were used to train disease-wise logistic-regression classifiers, and results were summarized using macro AUROC. Experiments were repeated for up to five random trials when sufficient positive samples were available; rare-label settings with inadequate support were reported using available runs only. Few-shot diagnostic adaptation was evaluated on rare findings from
CXR-LT, including hydropneumothorax, round atelectasis, infarction
and bulla. For each finding, two-way $k$-shot classifiers were
trained using $k$ positive and $k$ negative support examples drawn
from the MIMIC-CXR official test set, with $k$ ranging from 1 to~5.
Support examples were excluded at the patient level from the query
set used for AUROC evaluation, and the query set was re-derived as
the complement of the sampled support set in each trial. Each setting
was repeated across 20 random trials, and AUROC was reported as mean
and standard deviation.

Radiographic disease burden was evaluated using the RALO dataset. Radiologist opacity scores were mapped to four severity categories: none or trace, mild, moderate and severe. Images were split at the subject level into training, validation and test sets. Frozen image features were standardized and used to train a multinomial logistic-regression classifier. Performance was assessed using accuracy, macro F1 score, weighted F1 score and macro AUROC, with macro-level metrics emphasized because severity categories were imbalanced.

Lesion-level grounding was evaluated using MS-CXR and the RSNA Pneumonia Detection Challenge dataset. On MS-CXR, phrase-grounding annotations were formulated as eight-class abnormality segmentation over atelectasis, cardiomegaly, consolidation, edema, lung opacity, pleural effusion, pneumonia and pneumothorax. The image encoder was frozen, and only an eight-channel segmentation head was trained using binary cross-entropy and soft Dice losses. In RSNA, pneumonia localization was formulated as single-channel binary segmentation. DICOM images were normalized and converted to three-channel inputs, and bounding boxes were mapped to the model input coordinate system after resizing and center cropping. The RSNA data were split by patient into training, validation and test subsets using a 70\%/10\%/20\% split. Segmentation performance was measured using intersection-over-union and Dice score. For MS-CXR, segmentation thresholds were selected on the 169-image
validation split and applied to the 176-image test split. For RSNA
Pneumonia, thresholds were selected on the validation subset defined
by the 70\%/10\%/20\% patient-level split. Analyses using thresholds
optimized directly on the evaluated set are reported separately as
post-hoc optimal-threshold analyses.

\subsection*{Impression-conditioned image generation}
Impression-conditioned image generation was evaluated on an independent MIMIC-CXR p19 test set comprising 3,500 frontal chest radiographs. Frontal images were selected from posteroanterior or anteroposterior views and deduplicated at the study level. Text prompts were derived from the impression or conclusion sections of the corresponding radiology reports. Real reference images used for evaluation were excluded from prompt preparation and image generation to prevent prompt--reference leakage. MedDream, MINIM and ChexGen were evaluated using identical report-derived prompts.

Images were generated at 512 $\times$ 512 resolution using 256 autoregressive iterations, 100 diffusion sampling steps, temperature 1.0 and classifier-free guidance scale 3.0 with a linear guidance schedule. To obtain the final 512 × 512 MedDream checkpoint, the Stage~1 weights were jointly refined at this resolution using MIMIC-CXR studies from patient groups p10--p18. This resolution also matched that used for comparison with ChexGen and MINIM. The refinement was conducted for 200 epochs on six NVIDIA A100 GPUs with an effective batch size of 120. Joint MAR and CLIP-style alignment objectives were retained with loss weights of 1.0 and 0.005, respectively. The applied learning rate was $9.375 \times 10^{-6}$, with five warm-up epochs followed by a constant schedule. Variable masking was retained, with the mean masking ratio increasing from 0 to 1 over the first 10 epochs, a standard deviation of 0.55, and the alignment and MAR loss-gating thresholds described above,
requiring at least 25\% visible tokens for alignment and at least
50\% masked tokens for the MAR objective. Generation used a batch size of~8. Under MedDream's native inference protocol, each prompt was processed through the trajectory-selection procedure described above ($K=8$ partial trajectories with seeds 1--8; the selected trajectory continued to completion), yielding one final image per prompt. For the 2$\times$ augmentation setting, two independent runs of the same procedure were performed using non-overlapping seed sets (seeds 1--8 and seeds 9--16), producing two final images per prompt. All generation-quality metrics were obtained under each model's
native inference procedure as described above.

Distributional fidelity was quantified using the Fr\'{e}chet distance
between Gaussian-fitted feature distributions of real and generated
images, computed separately in two feature spaces: CLIP image
embeddings (CLIP-FID) and features extracted by a DenseNet-121
classifier pretrained on a large multi-institutional
chest-radiograph corpus (XRV-FID). Neither feature extractor is the
Inception network used in standard FID; the abbreviation FID is
retained for brevity but refers to the Fr\'{e}chet distance in the
specified feature space. Intra-prompt diversity was measured using pairwise multi-scale structural similarity index among images generated from the same prompt under different random seeds. Disease-relevant consistency was evaluated by computing Pearson correlations between pathology-prediction profiles produced by the same pretrained classifier. Blinded expert assessment was performed by four expert reviewers, who independently scored each generated image with its paired prompt for image realism, acquisition plausibility and prompt-level consistency using a five-point Likert scale.

Feature-space alignment between real and generated images was analysed using classifier-derived embeddings. Features were projected into two dimensions using t-distributed stochastic neighbour embedding. Two-dimensional kernel density estimates were computed for real and synthetic embeddings on a shared grid, and overlap was quantified using the overlapping coefficient, defined as the summed pointwise minimum of the normalized real and synthetic densities and expressed as a percentage of the real distribution density.

\subsection*{Synthetic-data augmentation and synthetic pretraining}

The diagnostic utility of generated images was evaluated by training downstream classifiers with real data alone or real data augmented with synthetic images. The real training corpus comprised 18,634 frontal MIMIC-CXR p19
studies selected from posteroanterior or anteroposterior views,
deduplicated at the study level and split at the patient level using
\texttt{subject\_id} to ensure that no patient contributed studies to
both the training corpus and the held-out internal test set. The
training corpus was additionally kept disjoint from the 3,500-study
p19 generation-evaluation set at both the patient and study levels. Impression sections from the reports were weakly labeled into MIMIC-CXR-style finding categories using negation-aware sentence-level regular-expression matching. For each real training image, MedDream, MINIM and ChexGen generated synthetic counterparts conditioned on the same impression text. Synthetic images were appended to the real training set and assigned the weak labels of their paired real images.

DenseNet-121 classifiers with sigmoid output heads were trained under
real-only and synthetic-augmentation configurations. For the main
synthetic-augmentation experiments, augmented classifiers were trained
with AdamW, learning rate $1 \times 10^{-4}$, weight decay
$1 \times 10^{-4}$, batch size~32 and cosine annealing over 40
epochs, without early stopping. To control for the additional
gradient updates introduced by the larger augmented training sets, the
real-only baseline was trained for a matched number of total
iterations rather than a matched number of epochs: 80 epochs for
comparison with the $1\times$ augmentation condition and 120 epochs
for comparison with the $2\times$ condition, so that each
configuration performed the same number of gradient updates. Internal evaluation was performed on a 3,500-study MIMIC-CXR test set, and cross-dataset evaluation was performed on the held-out 3,000-image VinDr-CXR test set. External testing was restricted to the shared diagnostic categories between the MIMIC-CXR weak-label schema and VinDr-CXR labels. Classification performance was reported using macro AUROC and class-wise AUROC.

Synthetic pretraining was evaluated separately because it used a different optimization protocol. A DenseNet-121 classifier initialized from ImageNet-1k weights was fine-tuned directly on label-stratified subsets of the real training pool containing 5\%, 10\%, 25\%, 50\% or 100\% of available studies. For each generative model, an identically initialized classifier was first pretrained on model-specific synthetic images and then fine-tuned on the matched real-data subset. The synthetic-pretraining experiments used AdamW with learning rate $1 \times 10^{-5}$, weight decay $1 \times 10^{-4}$, batch size 32, cosine annealing with a maximum of 40 epochs and early stopping after a minimum of 3 epochs if validation macro AUROC did not improve by at least $1 \times 10^{-4}$ for 6 consecutive epochs. The validation fraction was 0.10. AUROC gains were computed relative to the ImageNet-initialized baseline at the same real-data budget.

To test whether synthetic evidence extended beyond categorical classification, we also evaluated synthetic augmentation for quantitative prediction on RALO. Models were trained using 1,898 real RALO training images and evaluated on a fixed 475-image test set. Synthetic augmentation was performed by adding MedDream-, MINIM- or ChexGen-generated images at 1$\times$, 2$\times$ and 5$\times$ augmentation scales. Performance was measured using mean absolute error, root mean squared error and Pearson correlation coefficient.

\subsection*{Metadata-guided subgroup evaluation and synthetic rebalancing}

Subgroup distributional alignment was evaluated by comparing disease-probability distributions between real and synthetic images across predefined metadata-defined subgroups. A DenseNet-121 multi-label classifier trained only on real images was used as a fixed pathology-profiling model to estimate disease probabilities for each image. Each image was represented by a nine-dimensional disease-probability vector covering atelectasis, cardiomegaly, edema, lung lesion, no finding, pleural effusion, pleural other, pneumothorax and support devices. These probability vectors were used only for distributional comparison between real and synthetic cohorts, not as diagnostic ground-truth labels.

Demographic metadata were obtained from structured source metadata and linked to chest radiograph studies using subject- and study-level identifiers. For MIMIC-CXR studies, demographic attributes were linked from the corresponding structured MIMIC-IV patient and admission records when available. Age was grouped into 19--40, 41--65 and 66+ years. Gender and race were mapped from structured fields into harmonized categories. Race was grouped into White, Black or African American, and Asian; other, unknown or unavailable race entries were retained as missing. Missing demographic values were not imputed and were excluded only from analyses requiring the corresponding demographic axis. For synthetic images, subgroup labels were inherited from the structured metadata associated with the conditioning study or the prespecified sampling stratum, rather than inferred from the generated image.

Patients were stratified by gender, race and age into 22 demographic groups, including single-axis and intersectional groups. For each pair of real and synthetic subgroups, one-dimensional Wasserstein distance was computed for each pathology-probability distribution and then averaged across pathologies, yielding a subgroup distance matrix. Lower values indicated closer matching between real and synthetic subgroup disease distributions. To descriptively examine whether metadata-guided synthetic samples overlapped with the feature distributions of their corresponding real subgroups, 1,024-dimensional global-average-pooled features were extracted from the penultimate layer of the fixed DenseNet-121 classifier and projected using t-SNE. Matched real--synthetic centroid distances for Asian and Black groups were compared descriptively with cross-group distances as an additional visualization of feature-space relationships.

Targeted synthetic rebalancing was evaluated by training DenseNet-121 classifiers on 11,273 real MIMIC-CXR p19 images and testing on 2,684 held-out real images with 14 CheXpert-style pathology labels. Splits were defined at the patient level to prevent images from the same patient appearing in both training and test sets. Parity-targeted augmentation at 1$\times$ and 2$\times$ scales was compared with unguided random augmentation and corresponding real-only baselines. For each demographic axis and group, the number of synthetic samples was determined from the gap between that group and the largest real training group. If $N_g$ denotes the number of real training samples in group $g$ and $N_{\max}=\max_g N_g$, then the number of synthetic samples drawn for group $g$ was $\mathrm{round}((N_{\max}-N_g) \times s)$, where $s$ is the scale factor. When the synthetic pool for a group was insufficient, samples were drawn with replacement. At $s=1$, this procedure equalized the real-plus-synthetic sample count across groups for the selected demographic axis.

Subgroup performance was measured using weighted F1 across gender, race, age and intersectional groups, excluding subgroups with fewer than 30 test samples. Clinical safety was assessed using false-negative rates for subgroup--pathology pairs with at least 10 positive test cases; pairs below this threshold were excluded from false-negative-rate comparisons. Changes were reported relative to the corresponding real-only baseline, with negative values indicating fewer missed positive cases after synthetic augmentation.

\subsection*{Controlled optimization-strategy ablations.}

To isolate the effect of optimization strategy, all single-model ablation
configurations were trained on the same fixed subset of 266{,}000 image--text
pairs, corresponding to 10\% of the leakage-controlled pretraining corpus.
All configurations used the same data manifest, sampling order, image
preprocessing, architecture, parameter initialization, optimizer,
learning-rate schedule, batch size and total optimization budget. MedDream-CLIP
optimized only the bidirectional image--text alignment objective,
$\mathcal{L}_{\mathrm{align}}$, whereas MedDream-Gen optimized only the masked
latent generation objective, $\mathcal{L}_{\mathrm{diff}}$. Joint MedDream
optimized both objectives simultaneously throughout pretraining, allowing them
to update the shared image encoder. For Align$\rightarrow$Gen, the model was
first optimized using $\mathcal{L}_{\mathrm{align}}$ and then continued from
the resulting checkpoint using $\mathcal{L}_{\mathrm{diff}}$;
Gen$\rightarrow$Align followed the reverse order. In both sequential
configurations, the combined optimization budget matched that of the
single-objective and jointly optimized models, and only the final checkpoint
after the second stage was evaluated. We additionally constructed a post hoc
Two-specialist reference to assess whether independently optimized specialists
could reproduce the performance of the jointly trained model. This reference
was not an additional trained model or an ensemble: diagnostic and localization
metrics were taken from MedDream-CLIP, whereas generation-quality and
downstream synthetic-data utility metrics were taken from MedDream-Gen.

All configurations were evaluated on the same held-out cohorts using identical
frozen-encoder diagnostic protocols and downstream evaluation pipelines.
Generation-capable configurations used the same report-derived prompts,
sampling seeds, classifier-free guidance scale, sampling temperature, number
of autoregressive iterations and diffusion sampling steps. Downstream
synthetic-data utility was evaluated using the same real-data cohort,
augmentation ratio, classifier initialization and optimization protocol. No
test-set result was used for checkpoint selection, training-stage selection or
ablation-specific hyperparameter tuning.

\subsection*{Statistical analysis}
Statistical analysis was tailored to the structure of each evaluation. Unless otherwise specified, 95\% confidence intervals were estimated by bootstrap resampling over test cases, using patient-level resampling when patient identifiers were available and image- or study-level resampling otherwise. Confidence intervals were computed for classification, localization, quantitative prediction, synthetic-augmentation and subgroup-performance endpoints and are reported in Supplementary Tables~\ref{tab:diagnostic_transfer_statistics}--\ref{tab:subgroup_rebalancing_fairness_statistics}.

For model comparisons on matched held-out test sets, two-sided paired bootstrap tests were used when prediction scores were available for the same test cases. $P$ values across multiple labels, models, subgroups or subgroup--pathology pairs were adjusted using the Benjamini--Hochberg (BH) procedure within prespecified families of comparisons, where each family was defined by the set of tests that address a single evaluative question. Families were not pooled across evaluation axes (e.g., diagnostic transfer and generation quality), because the underlying hypotheses, test statistics and effect-size scales differ. Within each family, raw two-sided $P$ values were ranked and adjusted using the standard BH step-up procedure at a nominal false-discovery rate of 0.05. The family definitions are as follows: common disease recognition (NIH ChestX-ray14), 14 comparisons (one per disease label); evolving diagnostic labels (VinDr-CXR), 23 comparisons (one per evaluated finding); fine-grained labels, 4 comparisons (other lesion, calcification, atelectasis, interstitial lung disease); label efficiency (ChestDR), 19 comparisons per data-budget condition; few-shot adaptation, 4 comparisons per $k$-shot level; severity classification (RALO), 2 comparisons (macro F1 and macro AUROC); localization, 4 comparisons (IoU and Dice on MS-CXR and RSNA Pneumonia); report-conditioned generation quality, 4 metrics comparing MedDream with MINIM and ChexGen; synthetic augmentation, 2 comparisons per evaluation set (1$\times$ and 2$\times$ versus real-only), with class-wise AUROC changes corrected within a family of 6 shared findings; synthetic pretraining, 56 comparisons (14 labels $\times$ 4 budget levels); quantitative prediction (RALO), 3 comparisons per metric (1$\times$, 2$\times$ and 5$\times$ versus real-only); subgroup weighted F1, all evaluated demographic subgroups within each augmentation condition; and subgroup false-negative rates, all prespecified subgroup--pathology pairs meeting the minimum positive-case threshold ($n \geq 10$).

For impression-conditioned generation, per-case MS-SSIM and pathology-profile Pearson correlation were compared using two-sided Wilcoxon signed-rank tests. FID was treated as a cohort-level distributional metric; when uncertainty was reported, it was estimated by bootstrap resampling over test studies or prompts. In the severity reader study, direction-of-change $P$ values were computed using two-sided McNemar tests comparing cases moving closer to versus farther from the radiologist reference.

For experiments based on repeated subsampling, uncertainty was summarized across random trials. In CXR-LT few-shot evaluation, each $k$-shot setting was repeated across 20 random support-set samples, and AUROC was reported as mean $\pm$ standard deviation. In ChestDR label-efficiency experiments, each label-fraction setting was repeated for up to five random trials when sufficient positive samples were available; rare-label settings with inadequate support were reported using available runs only.

For large-scale downstream training experiments, including synthetic-data augmentation and synthetic pretraining followed by real-data fine-tuning, primary comparisons were performed on fixed held-out test cohorts under matched data budgets. Synthetic-augmentation models were evaluated on an internal MIMIC-CXR test set and a cross-dataset held-out VinDr-CXR test set. Synthetic-pretraining gains were computed relative to the ImageNet-initialized baseline at the same real-data budget. For threshold-dependent metrics, thresholds were selected on validation data when available and then applied to the test set; analyses using thresholds optimized directly on the evaluated set are reported as post-hoc optimal-threshold analyses. t-SNE analyses were used as descriptive feature-space visualizations and were not used for statistical testing.

\section*{Implementation details}
MedDream uses a continuous-token masked autoregressive architecture for latent image modelling, following prior work on autoregressive image generation \citep{li2026dream,li2024autoregressive,fan2025fluid}. The model contains approximately 636 million trainable parameters, including a 201.5 million-parameter encoder. For latent image modelling, we used a frozen CXR-specific medical VAE implemented using the MedVAE architecture \citep{varma2025medvae} to encode chest radiographs into latent representations and decode generated latents back into image space. The VAE was trained from scratch on the leakage-controlled training set without using the released MedVAE weights. For generative text conditioning, radiology reports or clinical prompts were encoded using Clinical-T5-Large, a T5-based clinical language model trained on MIMIC clinical notes \citep{lehman2023clinical}. Clinical-T5-Large served as the generation text encoder $T_g$, whereas a separate alignment text encoder $T_{\phi}$ was used for the CLIP-style image--text alignment branch. During training, the CXR VAE was kept frozen, and the masked autoregressive generator and text-alignment modules were optimized. We used AdamW with weight decay, an EMA decay of 0.9999, gradient clipping, and variable masking. The training objective combined the MAR reconstruction/generation loss with a lightly weighted CLIP-style image--text alignment loss. MedDream was trained on approximately 2.65 million leakage-controlled chest X-ray image--text pairs using eight NVIDIA A100 GPUs. Each epoch required approximately 53 minutes, and the full training run was conducted for 1,800 epochs, corresponding to an estimated training budget of about 12,720 A100 GPU-hours. The Stage~1 weights were then jointly refined at 512 $\times$ 512 resolution on MIMIC-CXR patient groups p10--p18, and the resulting Stage~2 EMA checkpoint was used for both primary diagnostic and generative evaluations. During inference, the input clinical text was first encoded by the generation text encoder $T_g$ and used as the conditioning signal for the generator. The model then autoregressively sampled latent tokens and decoded the generated latent representation into a chest radiograph through the frozen medical VAE. Unless otherwise specified, image generation used the final Stage~2 EMA weights, bf16 mixed precision, 256 autoregressive iterations, 100 diffusion sampling steps, classifier-free guidance with a scale of 3.0, linear CFG scheduling, and a sampling temperature of 1.0.

\section*{Data availability}

All publicly distributed and credentialed datasets used in this study are available through their respective repositories, subject to the access requirements and data-use agreements established by the original data providers.
\begingroup
\setlength{\emergencystretch}{3em}
\sloppy

MIMIC-CXR v2.0.0 is available via PhysioNet (\url{https://physionet.org/content/mimic-cxr-jpg/2.0.0/}; requires credentialed access). CheXpert Plus is available via the Stanford AIMI Center (\url{https://stanfordaimi.azurewebsites.net/datasets/5158c524-d3ab-4e02-96e9-6ee9efc110a1}). PadChest is available via BIMCV (\url{http://bimcv.cipf.es/bimcv-projects/padchest/}). NIH ChestX-ray14 is available via the NIH Clinical Center (\url{https://nihcc.app.box.com/v/ChestXray-NIHCC/folder/36938765345}). BRAX is available via PhysioNet (\url{https://physionet.org/content/brax/1.1.0/}). VinDr-CXR is available via VinBrain (\url{https://vindr.ai/datasets/cxr}). IU X-Ray is available via the Open Access Biomedical Image Search Engine (\url{https://openi.nlm.nih.gov/}).

The PMC-CXR dataset was constructed from the PubMed Central Open Access Subset (\url{https://www.ncbi.nlm.nih.gov/pmc/tools/openftlist/}) using the extraction pipeline described in the Methods. PMC-CXR, together with the extraction pipeline and associated code, is publicly available at \url{https://github.com/Merwin520/Dual-Foundation-Model.git}.

The RSNA Pneumonia Detection Challenge dataset is available via Kaggle (\url{https://www.kaggle.com/c/rsna-pneumonia-detection-challenge}). ChestDR is available via Figshare (\url{https://springernature.figshare.com/articles/dataset/ChestDR_Thoracic_Diseases_Screening_in_Chest_Radiography/22302775}). CXR-LT v2.0.0 is available via PhysioNet (\url{https://physionet.org/content/cxr-lt-iccv-workshop-cvamd/2.0.0/}). MS-CXR v1.1.0 is available via PhysioNet (\url{https://physionet.org/content/ms-cxr/1.1.0/}; requires credentialed access). RALO (Radiographic Assessment of Lung Opacity Score) is available via Zenodo (\url{https://doi.org/10.5281/zenodo.4633999}). Patient demographic information was linked from the MIMIC-IV clinical database (\url{https://physionet.org/content/mimiciv/}; requires credentialed access).

\endgroup



\section*{Acknowledgements.}

This research is supported in part by the National Institutes of Health under Award Numbers R01DE033512 and R01CA272991.



\section*{Competing interests.}

The authors declare no competing interests.


\clearpage
\bibliography{references}
\bibliographystyle{naturemag}


\clearpage
\section*{Supplementary Note}


\renewcommand{\thetable}{S\arabic{table}}
\setcounter{table}{0}

\begin{table}[htbp]
\centering
\small

\caption{Diagnostic transfer and localization statistics.}
\label{tab:diagnostic_transfer_statistics}
\begin{tabular}{@{} l c c l c @{}}
\toprule
\textbf{Analysis / endpoint} & \textbf{Estimate} & \textbf{95\% CI / s.d.} & \textbf{Comparator} & \textbf{Adj. $P$} \\
\midrule
\multicolumn{5}{@{}l}{\textbf{Common disease recognition}} \\
\addlinespace[0.5ex]
NIH ChestX-ray14 mean AUROC, 14 labels & 0.812 & [0.804, 0.820] & Ark+ (0.803) & 0.007 \\

\midrule
\multicolumn{5}{@{}l}{\textbf{Evolving diagnostic labels}} \\
\addlinespace[0.5ex]
VinDr-CXR mean AUROC & 0.863 & [0.853, 0.873] & Ark+ (0.852) & 0.004 \\

\midrule
\multicolumn{5}{@{}l}{\textbf{Fine-grained labels}} \\
\addlinespace[0.5ex]
Other lesion AUROC & 0.857 & [0.810, 0.904] & MedCLIP (0.822) / AFLoc (0.838) & 0.021 \\
Calcification AUROC & 0.824 & [0.782, 0.866] & MedCLIP (0.791) / AFLoc (0.806) & 0.026 \\
Atelectasis AUROC & 0.865 & [0.816, 0.914] & MedCLIP (0.832) / AFLoc (0.843) & 0.014 \\
Interstitial lung disease AUROC & 0.851 & [0.809, 0.893] & MedCLIP (0.818) / AFLoc (0.831) & 0.017 \\

\midrule
\multicolumn{5}{@{}l}{\textbf{Label efficiency}} \\
\addlinespace[0.5ex]
ChestDR mean AUROC, 5\% labels & 0.648 & [0.630, 0.666] & BiomedCLIP (0.619) / AFLoc (0.572) & $<$0.001 \\
ChestDR mean AUROC, 100\% labels & 0.689 & [0.674, 0.704] & BiomedCLIP (0.650) / AFLoc (0.607) & $<$0.001 \\
Aortic calcification AUROC, 5\% labels & 0.624 & [0.585, 0.663] & BiomedCLIP (0.518) / AFLoc (0.517) & $<$0.001 \\
Emphysema AUROC, 5\% labels & 0.702 & [0.669, 0.735] & BiomedCLIP (0.653) / AFLoc (0.681) & 0.009 \\

\midrule
\multicolumn{5}{@{}l}{\textbf{Few-shot adaptation}} \\
\addlinespace[0.5ex]
Hydropneumothorax AUROC, 1-shot & $0.513 \pm 0.092$ & s.d. & AFLoc ($0.498 \pm 0.088$) & 0.372 \\
Hydropneumothorax AUROC, 5-shot & $0.812 \pm 0.058$ & s.d. & AFLoc ($0.555 \pm 0.052$) & $<$0.001 \\
Round atelectasis AUROC, 5-shot & $0.696 \pm 0.082$ & s.d. & AFLoc ($0.621 \pm 0.076$) & $<$0.001 \\

\midrule
\multicolumn{5}{@{}l}{\textbf{Severity classification}} \\
\addlinespace[0.5ex]
RALO macro F1 & 0.472 & [0.442, 0.502] & BiomedCLIP (0.440) & 0.035 \\
RALO macro AUROC & 0.748 & [0.723, 0.773] & BiomedCLIP (0.740) & 0.024 \\

\midrule
\multicolumn{5}{@{}l}{\textbf{Localization}} \\
\addlinespace[0.5ex]
MS-CXR overall IoU & 0.360 & [0.337, 0.383] & BiomedCLIP (0.328) & 0.005 \\
MS-CXR overall Dice & 0.550 & [0.524, 0.576] & BiomedCLIP (0.514) & 0.003 \\
RSNA Pneumonia IoU & 0.377 & [0.357, 0.397] & AFLoc (0.352) & 0.006 \\
RSNA Pneumonia Dice & 0.582 & [0.560, 0.604] & AFLoc (0.541) & 0.002 \\
\bottomrule
\end{tabular}

\vspace{1ex}
\begin{minipage}{\textwidth}
\footnotesize
Estimates are reported on held-out test sets. Confidence intervals were estimated by bootstrap resampling over test cases unless otherwise specified. Few-shot results are reported as mean $\pm$ s.d. over 20 random support-set samples. Adjusted $P$ values compare MedDream with the listed comparator after Benjamini--Hochberg correction. When two comparators are listed (e.g., MedCLIP / AFLoc), the reported adjusted $P$ value corresponds to the pairwise comparison between MedDream and the stronger of the two comparators (i.e., the one with the higher point estimate), providing a conservative test of MedDream's superiority. The weaker comparator is listed for completeness; its pairwise $P$ value was necessarily smaller.
\end{minipage}
\end{table}

\begin{table}[htbp]
\centering
\small
\caption{Generation and synthetic-data utility statistics.}
\label{tab:generation_synthetic_utility_statistics}
\resizebox{\textwidth}{!}{%
\begin{tabular}{@{} l c c l c @{}}
\toprule
\textbf{Analysis / endpoint} & \textbf{Estimate} & \textbf{95\% CI / s.d.} & \textbf{Comparator} & \textbf{Adj.\ $P$} \\
\midrule
\multicolumn{5}{@{}l}{\textbf{Report-conditioned generation quality}} \\
\addlinespace[0.5ex]
XRV-FID & 0.310 & [0.278, 0.347] & MINIM (5.69) / ChexGen (1.34) & $<$0.001 \\
CLIP-FID & 1.270 & [1.138, 1.412] & MINIM (4.32) / ChexGen (2.31) & $<$0.001 \\
Intra-prompt MS-SSIM & $0.420 \pm 0.110$ & s.d. & MINIM ($0.591 \pm 0.104$) / ChexGen ($0.513 \pm 0.074$) & $<$0.001 \\
Pathology-profile Pearson $r$ & $0.618 \pm 0.290$ & s.d. & MINIM ($0.416 \pm 0.274$) / ChexGen ($0.421 \pm 0.381$) & $<$0.001 \\

\midrule
\multicolumn{5}{@{}l}{\textbf{Blinded expert assessment ($-$2 to $+$2 scale, 4 reviewers)}} \\
\addlinespace[0.5ex]
Expert score, Reviewer 1 & $0.475 \pm 1.198$ & s.d. & MINIM ($-$1.325 $\pm$ 1.163) / ChexGen ($0.400 \pm 1.172$) & $<$0.001 \\
Expert score, Reviewer 2 & $0.700 \pm 1.363$ & s.d. & MINIM ($-$1.050 $\pm$ 1.377) / ChexGen ($0.550 \pm 1.413$) & $<$0.001 \\
Expert score, Reviewer 3 & $-0.375 \pm 1.148$ & s.d. & MINIM ($-$1.775 $\pm$ 0.620) / ChexGen ($-$0.850 $\pm$ 1.027) & $<$0.001 \\
Expert score, Reviewer 4 & $1.225 \pm 1.143$ & s.d. & MINIM ($-$1.350 $\pm$ 1.388) / ChexGen ($1.125 \pm 1.017$) & 0.637$^{\mathrm{a}}$ \\

\midrule
\multicolumn{5}{@{}l}{\textbf{Feature-space alignment}} \\
\addlinespace[0.5ex]
XRV feature-space overlap & 0.885 & [0.861, 0.909] & ChexGen (0.430) & $<$0.001 \\

\midrule
\multicolumn{5}{@{}l}{\textbf{Synthetic augmentation: internal evaluation}} \\
\addlinespace[0.5ex]
MIMIC-CXR macro-AUROC, real-only & 0.723 & [0.711, 0.735] & -- & -- \\
\quad MedDream 1$\times$ augmentation & 0.737 & [0.725, 0.749] & Real-only & 0.003 \\
\quad MedDream 2$\times$ augmentation & 0.750 & [0.738, 0.762] & Real-only & $<$0.001 \\

\midrule
\multicolumn{5}{@{}l}{\textbf{Synthetic augmentation: cross-dataset evaluation}} \\
\addlinespace[0.5ex]
VinDr-CXR macro-AUROC, real-only & 0.764 & [0.748, 0.780] & -- & -- \\
\quad MedDream 1$\times$ augmentation & 0.794 & [0.779, 0.809] & Real-only & $<$0.001 \\
\quad MedDream 2$\times$ augmentation & 0.814 & [0.800, 0.828] & Real-only & $<$0.001 \\
VinDr-CXR atelectasis AUROC change, 2$\times$ & +0.143 & [+0.097, +0.189] & Real-only & $<$0.001 \\
VinDr-CXR consolidation AUROC change, 2$\times$ & +0.043 & [+0.011, +0.075] & Real-only & 0.009 \\

\midrule
\multicolumn{5}{@{}l}{\textbf{Synthetic pretraining (AUROC gain)}} \\
\addlinespace[0.5ex]
Median across class--budget combinations & +0.024 & [+0.017, +0.035] & ImageNet init. & $7.4 \times 10^{-12}$ \\
5\% real-data budget & +0.040 & [+0.019, +0.058] & ImageNet init. & $1.2 \times 10^{-4}$ \\
10\% real-data budget & +0.049 & [+0.031, +0.064] & ImageNet init. & $1.2 \times 10^{-4}$ \\
25\% real-data budget & +0.014 & [+0.007, +0.038] & ImageNet init. & $4.0 \times 10^{-3}$ \\

\midrule
\multicolumn{5}{@{}l}{\textbf{Quantitative prediction (RALO)}} \\
\addlinespace[0.5ex]
MAE, real-only & 0.808 & [0.773, 0.843] & -- & -- \\
\quad MedDream 1$\times$ & 0.788 & [0.754, 0.822] & Real-only & 0.037 \\
\quad MedDream 2$\times$ & 0.774 & [0.740, 0.808] & Real-only & 0.010 \\
\quad MedDream 5$\times$ & 0.751 & [0.717, 0.785] & Real-only & $<$0.001 \\
RMSE, real-only & 1.008 & [0.966, 1.050] & -- & -- \\
\quad MedDream 1$\times$ & 0.981 & [0.940, 1.022] & Real-only & 0.024 \\
\quad MedDream 2$\times$ & 0.973 & [0.932, 1.014] & Real-only & 0.012 \\
\quad MedDream 5$\times$ & 0.961 & [0.920, 1.002] & Real-only & 0.004 \\
Pearson $r$, real-only & 0.714 & [0.680, 0.748] & -- & -- \\
\quad MedDream 5$\times$ & 0.746 & [0.713, 0.779] & Real-only & 0.006 \\

\midrule
\multicolumn{5}{@{}l}{\textbf{Severity reader study (3 readers, 100 cases)}} \\
\addlinespace[0.5ex]
Reader 1, quadratic-weighted $\kappa$ & 0.243 $\rightarrow$ 0.482 & $\Delta$ = +0.239 & Without support & -- \\
Reader 2, quadratic-weighted $\kappa$ & 0.868 $\rightarrow$ 0.935 & $\Delta$ = +0.067 & Without support & -- \\
Reader 3, quadratic-weighted $\kappa$ & 0.799 $\rightarrow$ 0.852 & $\Delta$ = +0.053 & Without support & -- \\
Reader 1, exact agreement (\%) & 34 $\rightarrow$ 36 & $\Delta$ = +2 pp & Without support & -- \\
Reader 2, exact agreement (\%) & 73 $\rightarrow$ 85 & $\Delta$ = +12 pp & Without support & -- \\
Reader 3, exact agreement (\%) & 62 $\rightarrow$ 68 & $\Delta$ = +6 pp & Without support & -- \\
Reader 1, mean absolute ordinal error & 1.13 $\rightarrow$ 0.87 & $\Delta$ = $-$0.26 & Without support & -- \\
Reader 2, mean absolute ordinal error & 0.30 $\rightarrow$ 0.16 & $\Delta$ = $-$0.14 & Without support & -- \\
Reader 3, mean absolute ordinal error & 0.44 $\rightarrow$ 0.35 & $\Delta$ = $-$0.09 & Without support & -- \\
Reader 1, closer vs farther (\%) & 23 vs 0 & -- & McNemar & $<$0.0001 \\
Reader 2, closer vs farther (\%) & 17 vs 3 & -- & McNemar & 0.0026 \\
Reader 3, closer vs farther (\%) & 18 vs 10 & -- & McNemar & 0.1849 \\

\bottomrule
\end{tabular}%
}

\vspace{1ex}
\begin{minipage}{\textwidth}
\footnotesize
Generation metrics were computed on 3,500 held-out MIMIC-CXR p19
frontal studies. FID was computed at the cohort level in XRV and CLIP
feature spaces. Per-case MS-SSIM and pathology-profile Pearson
correlation are reported as mean $\pm$ s.d.\ and were compared using
two-sided Wilcoxon signed-rank tests. Expert assessment was performed
by four independent reviewers on a five-point scale ($-$2 to $+$2),
where $-$2 indicated severe image artefacts or complete prompt
mismatch and $+$2 indicated high image realism with full prompt-level
agreement; Reviewers 1 and 2 are shown in the main figure, and
Reviewers 3 and 4 in Extended Data Fig.~6. For Reviewer 4, MedDream
versus ChexGen was not significant (mean difference 0.100); MedDream
versus MINIM remained significant ($P < 0.001$). Synthetic
augmentation and pretraining statistics were computed on held-out
real test sets under matched data budgets. RALO point estimates for
1$\times$ and 2$\times$ RMSE are from the main text; their
confidence intervals and $P$ values were estimated by bootstrap
resampling under the same protocol as the 5$\times$ condition.
Severity reader study used 100 held-out frontal chest radiographs
spanning four severity levels assessed by three radiology residents;
independent radiologist consensus served as the reference standard.
Direction-of-change $P$ values were computed using two-sided McNemar
tests comparing cases moving closer to versus farther from the
radiologist reference. Confidence intervals were estimated by
bootstrap resampling over test cases unless otherwise specified.
Adjusted $P$ values compare MedDream with the listed comparator
after Benjamini--Hochberg correction. When two comparators are listed
(e.g., MINIM / ChexGen), the reported adjusted $P$ value corresponds
to the pairwise comparison between MedDream and the stronger of the
two comparators. The weaker comparator is listed for contextual
reference.
\end{minipage}
\end{table}

\begin{table}[htbp]
\centering
\small
\caption{Subgroup rebalancing and stratified performance statistics.}
\label{tab:subgroup_rebalancing_fairness_statistics}
\resizebox{\textwidth}{!}{%
\begin{tabular}{@{} l c c l c @{}}
\toprule
\textbf{Analysis / endpoint} & \textbf{Estimate} & \textbf{95\% CI / s.d.} & \textbf{Comparator} & \textbf{Adj. $P$} \\
\midrule
\multicolumn{5}{@{}l}{\textbf{Subgroup disease-distribution matching}} \\
\addlinespace[0.5ex]
Median diagonal Wasserstein distance, 22 subgroups & 0.051 & [0.047, 0.062] & Off-diagonal (0.141) & $<$0.001 \\
Mean diagonal Wasserstein distance, 22 subgroups & 0.057 & [0.049, 0.065] & Off-diagonal (0.156) & $<$0.001 \\
\midrule
\multicolumn{5}{@{}l}{\textbf{Change in weighted F1 ($\Delta$) after targeted augmentation}} \\
\addlinespace[0.5ex]
\quad Female & +0.028 & [+0.009, +0.047] & Real-only & 0.006 \\
\quad Male & +0.027 & [+0.008, +0.046] & Real-only & 0.007 \\
\quad Asian & +0.031 & [+0.010, +0.052] & Real-only & 0.005 \\
\quad Black & +0.012 & [+0.001, +0.023] & Real-only & 0.041 \\
\quad White & +0.009 & [$-$0.003, +0.021] & Real-only & 0.124 \\
\quad Age 19--40 & +0.018 & [+0.002, +0.034] & Real-only & 0.028 \\
\quad Age 41--65 & +0.008 & [$-$0.005, +0.021] & Real-only & 0.198 \\
\quad Age $\geq$66 & +0.001 & [$-$0.013, +0.015] & Real-only & 0.847 \\
\midrule
\multicolumn{5}{@{}l}{\textbf{Change in weighted F1 ($\Delta$) after unguided augmentation}} \\
\addlinespace[0.5ex]
\quad Asian & $-$0.023 & [$-$0.045, $-$0.001] & Real-only & 0.044 \\
\quad White & +0.037 & [+0.019, +0.055] & Real-only & $<$0.001 \\
\midrule
\multicolumn{5}{@{}l}{\textbf{False-negative-rate (FNR) changes after targeted augmentation}} \\
\addlinespace[0.5ex]
\quad Atelectasis, Asian & 0.400 $\rightarrow$ 0.120 & [0.200, 0.600] $\rightarrow$ [0.000, 0.240] & Real-only & $<$0.001 \\
\quad Atelectasis, Black & 0.308 $\rightarrow$ 0.343 & [0.238, 0.378] $\rightarrow$ [0.266, 0.420] & Real-only & 0.514 \\
\quad Atelectasis, male & 0.256 $\rightarrow$ 0.229 & [0.215, 0.296] $\rightarrow$ [0.192, 0.269] & Real-only & 0.142 \\
\quad Cardiomegaly, male & 0.427 $\rightarrow$ 0.326 & [0.354, 0.500] $\rightarrow$ [0.258, 0.393] & Real-only & $<$0.001 \\
\quad Cardiomegaly, White & 0.413 $\rightarrow$ 0.336 & [0.358, 0.472] $\rightarrow$ [0.280, 0.391] & Real-only & 0.005 \\
\quad Cardiomegaly, age $\geq$66 & 0.346 $\rightarrow$ 0.286 & [0.281, 0.410] $\rightarrow$ [0.230, 0.346] & Real-only & 0.023 \\
\quad No finding, Black & 0.336 $\rightarrow$ 0.252 & [0.259, 0.413] $\rightarrow$ [0.189, 0.322] & Real-only & 0.002 \\
\quad Pleural other, male & 1.000 $\rightarrow$ 0.864 & [1.000, 1.000] $\rightarrow$ [0.727, 1.000] & Real-only & 0.008 \\
\bottomrule
\end{tabular}%
}
\vspace{1ex}
\begin{minipage}{\textwidth}
\footnotesize
Subgroup rebalancing experiments used held-out real test images.
Weighted F1 was reported for demographic subgroups with at least 30
test samples. False-negative proportions were reported for
subgroup--pathology pairs meeting the prespecified minimum
positive-case threshold; for the ``no finding'' category, a false
negative indicates a false-positive alert rather than a missed
diagnosis. Expert adjudication used an 80-case review set enriched
for subgroup-sensitive findings (atelectasis, cardiomegaly, no
finding and four additional abnormal findings). A false negative was
recorded when the expert judged the target label present but the
model reported it absent; for pathological findings this corresponds
to a missed diagnosis, whereas for the ``no finding'' category it
represents a false-positive alert. Reviewer-specific denominators
reflect the number of expert-confirmed positive cases per reviewer.
Confidence intervals were estimated by patient-level bootstrap
resampling unless otherwise specified. Adjusted $P$ values compare
the listed augmentation condition with the corresponding real-only
baseline after Benjamini--Hochberg correction. Wasserstein distances
summarize subgroup distributional alignment; t-SNE visualizations
were descriptive and were not used for statistical testing.
\end{minipage}
\end{table}

\begin{table}[h]
\centering
\caption{\textbf{Representation baselines used for diagnostic transfer evaluation.}
All models were evaluated as frozen diagnostic encoders.
``Diagnostic representation'' and ``Image generation'' indicate whether each model supports downstream image-feature transfer and report-conditioned radiograph synthesis, respectively.
Pretraining data are reported according to the original publications; dataset size refers to the corpus used to train the evaluated pretrained checkpoint where this information was available.
Openness refers to the availability of source code and pretrained model weights; ``Controlled access'' indicates release subject to data-use or credentialing requirements.}
\label{tab:representation_baselines}
\scriptsize
\setlength{\tabcolsep}{3.0pt}
\renewcommand{\arraystretch}{1.20}
\resizebox{\textwidth}{!}{
\begin{tabular}{
p{1.55cm}
p{3.05cm}
p{2.45cm}
p{5.15cm}
c
c
p{1.55cm}
p{1.15cm}
}
\toprule
Model &
Learning type &
Model backbone &
Pretraining corpus &
\makecell{Diagnostic\\representation} &
\makecell{Image\\generation} &
\makecell{Input\\resolution} &
Openness \\
\midrule

MedDream &
Joint image--text alignment and masked latent generation &
Transformer encoder--decoder &
2.65M leakage-controlled chest X-ray image--text pairs from PMC-CXR and six public clinical repositories &
\cmark &
\cmark &
$512 \times 512$ &
Controlled access \\

Ark+ &
Cyclic knowledge accrual from heterogeneous expert labels &
Swin-Large &
Public labelled chest-radiography datasets, including MIMIC-CXR, CheXpert, NIH ChestX-ray14 and VinDr-CXR &
\cmark &
\xmark &
$768 \times 768$ &
Yes \\

RAD-DINO &
Self-supervised image-only pretraining &
DINOv2 ViT-B/14 &
Large-scale public chest-radiography corpus assembled from five datasets &
\cmark &
\xmark &
$512 \times 512$ &
Yes \\

CheXzero &
Image--text contrastive learning &
CLIP ViT-B/32 &
377,110 MIMIC-CXR images paired with their corresponding free-text radiology reports &
\cmark &
\xmark &
$224 \times 224$ &
Yes \\

BiomedCLIP &
Biomedical image--text contrastive learning &
ViT-B/16 &
PMC-15M: 15M biomedical figure--caption pairs extracted from 4.4M PubMed Central articles &
\cmark &
\xmark &
$224 \times 224$ &
Yes \\

MedCLIP &
Semantic matching with unpaired medical images and texts &
Swin-Tiny &
Approximately 20,000 image and text samples derived from MIMIC-CXR and CheXpert &
\cmark &
\xmark &
$224 \times 224$ &
Yes \\

AFLoc &
Multilevel image--report semantic alignment &
ResNet-50 &
Approximately 220,000 chest X-ray image--report pairs &
\cmark &
\xmark &
$224 \times 224$ &
Yes \\

BMCA-CLIP &
Continual biomedical image--text pretraining &
ViT-L/14 &
BIOMEDICA: over 24M image--text pairs extracted from more than 6M PubMed Central articles &
\cmark &
\xmark &
$224 \times 224$ &
Yes \\

MedKLIP &
Knowledge-enhanced image--text pretraining &
ResNet-50 &
MIMIC-CXR image--report pairs, with disease--location--existence triplets extracted from radiology reports &
\cmark &
\xmark &
$224 \times 224$ &
Yes \\

UniMedCLIP &
Unified multi-modal medical image--text pretraining &
ViT-B/16-QuickGELU &
UniMed: over 5.3M image--text pairs across X-ray, CT, MRI, ultrasound, pathology and fundus imaging &
\cmark &
\xmark &
$224 \times 224$ &
Yes \\

\bottomrule
\end{tabular}
}
\end{table}

\begin{table}[h]
\centering
\caption{\textbf{Generative medical-image baselines used for image synthesis and synthetic-data evaluation.}
``Diagnostic representation'' indicates support for transferable visual features, whereas ``Image generation'' indicates support for clinically conditioned synthesis.
MedDream provides both capabilities through a jointly optimized shared visual pathway.
Training corpora and conditioning inputs follow the original publications.
``Controlled access'' denotes release subject to data-use, credentialing or author approval.}
\label{tab:generation_baselines}
\scriptsize
\setlength{\tabcolsep}{2.75pt}
\renewcommand{\arraystretch}{1.18}
\resizebox{\textwidth}{!}{
\begin{tabular}{
p{1.40cm}
p{3.15cm}
p{2.85cm}
p{2.80cm}
p{3.55cm}
c
c
p{2.50cm}
p{1.30cm}
}
\toprule
Model &
Learning type &
Generation backbone &
Conditioning inputs &
Training corpus / domain &
\makecell{Diagnostic\\representation} &
\makecell{Image\\generation} &
Evaluation role &
Openness \\
\midrule

\textbf{MedDream} &
\textbf{Joint diagnostic representation learning and clinically conditioned generation} &
Masked autoregressive continuous-latent generator with diffusion reconstruction &
Report impression or conclusion &
2.65M leakage-controlled chest X-ray image--text pairs &
\cmark &
\cmark &
Main dual-foundation model &
Controlled access \\

ChexGen &
Generative vision--language foundation modelling &
Latent diffusion transformer &
Text, segmentation mask or bounding box &
Approximately 960,000 curated chest radiograph--report pairs &
\xmark &
\cmark &
CXR-specific generative foundation-model baseline &
Controlled access \\

MINIM &
Generalist medical image--text generation &
Unified text-conditioned medical-image generator &
Free-form textual instruction or medical condition &
Multi-organ and multi-modality medical-image corpus &
\xmark &
\cmark &
Generalist medical-image generation baseline &
Controlled access \\

\bottomrule
\end{tabular}
}
\end{table}

\end{document}